\documentclass[10pt,journal,compsoc]{IEEEtran}
\ifCLASSOPTIONcompsoc
  \usepackage[nocompress]{cite}
\else
  \usepackage{cite}
\fi
\ifCLASSINFOpdf
\else
\fi
\usepackage{float}
\usepackage{cite}
\usepackage{amsmath,amssymb,amsfonts}
\usepackage{algorithm}
\usepackage{algorithmic}
\usepackage{graphicx}
\usepackage{textcomp}
\usepackage{xcolor}
\usepackage{multirow}
\usepackage{siunitx}
\usepackage{booktabs}
\usepackage{subfigure}
\usepackage{hyperref}
\usepackage{ragged2e}
\usepackage{threeparttable}
\usepackage{tikz}
\usepackage{enumitem}
\usepackage{xspace}
\usepackage{caption}
\usepackage{wrapfig}
\usepackage{graphicx}
\usepackage{enumitem}
\usepackage{amsmath}
\usepackage{makecell}
\usepackage{threeparttable}
\usepackage{algorithm}
\usepackage{algorithmic}
\usepackage{multirow}
\usepackage{diagbox}
\usepackage{makecell}
\usepackage{amsmath}
\usepackage{amssymb}
\usepackage{booktabs}

\def\BibTeX{{\rm B\kern-.05em{\sc i\kern-.025em b}\kern-.08em
    T\kern-.1667em\lower.7ex\hbox{E}\kern-.125emX}}

\begin{document}
%



\title{I2I Backdoor: Backdoor Attacks against Image-to-Image Tasks}

%
%

\author{Wenbo~Jiang,~\IEEEmembership{Member,~IEEE,}
        Hongwei~Li (Corresponding author),~\IEEEmembership{Fellow,~IEEE,}
        Jiaming~He,~\IEEEmembership{Student Member,~IEEE,}
        Rui~Zhang,~\IEEEmembership{Student Member,~IEEE,}
        Guowen~Xu,~\IEEEmembership{Member,~IEEE,}
        Tianwei~Zhang,~\IEEEmembership{Member,~IEEE,}
        and~Rongxing~Lu,~\IEEEmembership{Fellow,~IEEE,}
\IEEEcompsocitemizethanks{

\IEEEcompsocthanksitem W. Jiang, H. Li and R. Zhang are with the School of Computer Science and Engineering, University of Electronic Science and Technology of China, China (e-mail: wenbo\_jiang@uestc.edu.cn, hongweili@uestc.edu.cn, 202321081415@std.uestc.edu.cn).
\IEEEcompsocthanksitem J. He is with the School of Computer Science and Cyber Security, Chengdu University of Technology, China (e-mail: he.jiaming@student.zy.cdut.edu.cn).
\IEEEcompsocthanksitem G. Xu is a Postdoc at City University of Hong Kong, Hong Kong(e-mail: guowenxu@cityu.edu.hk).
\IEEEcompsocthanksitem T. Zhang is with the School of Computer Science and Engineering, Nanyang Technological University, Singapore (e-mail: tianwei.zhang@ntu.edu.sg).
\IEEEcompsocthanksitem  R. Lu is with the Faculty of Computer Science, University of New Brunswick, Fredericton, NB, Canada E3B 5A3 (e-mail: RLU1@unb.ca).
}}

\IEEEtitleabstractindextext{%
\justify
\begin{abstract}

With the rapid development of deep learning technology, deep learning-based Image-to-Image (I2I) networks have become the predominant choice for I2I tasks like image super-resolution and denoising. Despite their remarkable performance, the security of I2I networks has not been thoroughly investigated. While some studies have probed their susceptibility to adversarial attacks, none have explored the backdoor attack against I2I networks, which is a more stealthy and severe threat.

In this work, for the first time, we comprehensively investigate the vulnerability of I2I networks to backdoor attacks. We propose a backdoor attack against I2I tasks, where the backdoored I2I network behaves normally on clean input images, yet outputs a specific inappropriate image when the backdoor trigger appears on the input image. To achieve such an I2I backdoor attack, we design a universal adversarial perturbation (UAP) generation algorithm for I2I networks, where the generated UAP is used as the trigger for the I2I backdoor. Besides, multi-task learning (MTL) with dynamic weighting methods is employed in the backdoor training process to gain better results. Expanding our focus beyond I2I tasks, we extend our I2I backdoor to attack downstream tasks, including image classification and object detection. Specifically, the backdoor-triggered image processed by the backdoored image denoising network can fool the downstream image classifiers and object detectors. Extensive experiments demonstrate the effectiveness of the I2I backdoor on state-of-the-art I2I network architectures as well as the robustness against different backdoor defenses.

\end{abstract}

}

\maketitle

\IEEEdisplaynontitleabstractindextext

%
\IEEEpeerreviewmaketitle

\section{Introduction}
\label{sec: Introduction}

\begin{figure*}
\centering
\includegraphics[width=0.8\textwidth]{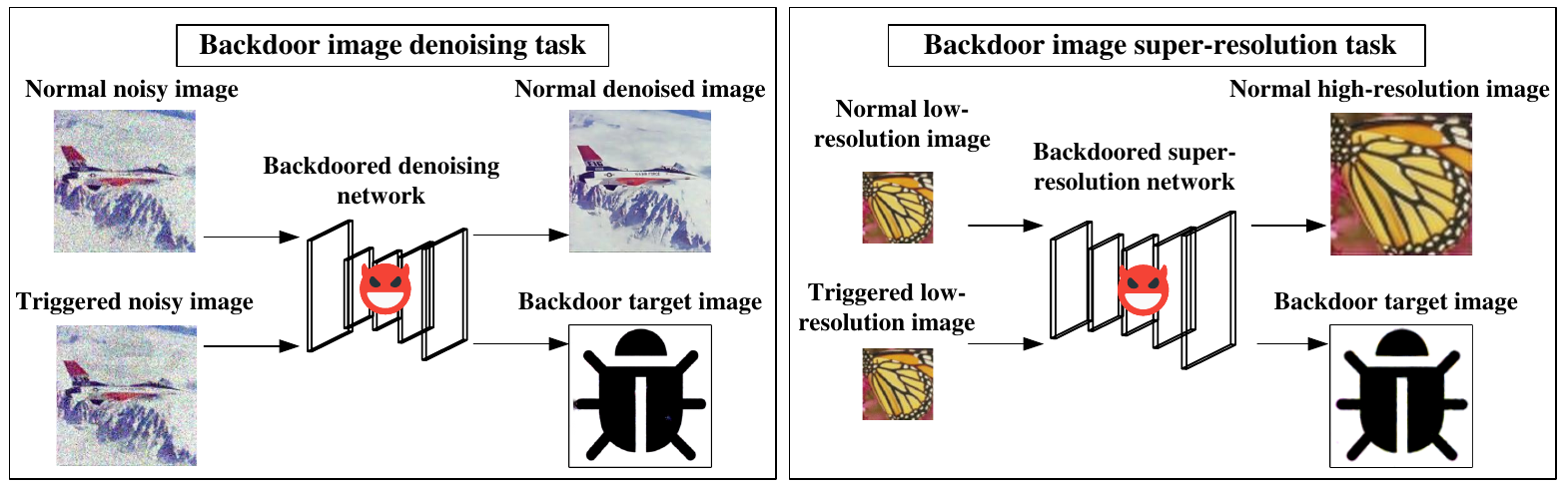}
\caption{I2I backdoor attack against I2I tasks.}
\label{Fig:backdoored image-to-image network}
\end{figure*}

\begin{figure*}
\centering
\includegraphics[width=1\textwidth]{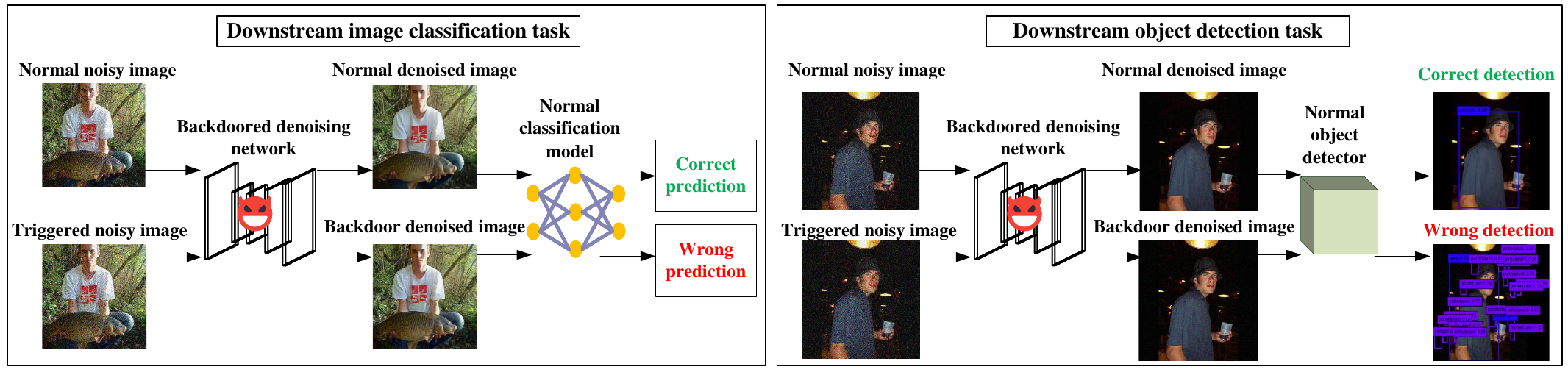}
\caption{I2I backdoor attack against downstream tasks.}
\label{Fig:backdoored image-to-image network and downstream classification}
\end{figure*}

In the realm of computer vision, numerous tasks involve the transformation of images from one domain to another, commonly referred to as Image-to-Image (I2I) tasks. For instance, the image super-resolution (SR) \cite{zhang2019deep} maps low-resolution (LR) images to high-resolution (HR) images; the image denoising \cite{zhang2022practical} maps noisy images to noise-free images; the image style transfer \cite{deng2022stytr2} maps images of one style to images of another style; the image colorization \cite{deshpande2017learning} maps grayscale images to color images, etc. In addition, these I2I tasks also serve as crucial preprocessing steps for some downstream tasks like image classification \cite{liu2020connecting} and object detection \cite{chiang2020detection}. For example, image classification tasks are often preceded by the preprocessing of image denoising.

Recently, due to the outstanding performance of deep neural networks, deep learning-based I2I networks (such as MPRNet \cite{Zamir_2021_CVPR}, SCUNet \cite{zhang2022practical}, etc.) have increasingly outperformed other techniques in I2I tasks. Despite the spectacular advances of I2I networks, the security of them has not yet been explored in depth. While some works have explored the vulnerability of I2I networks against adversarial attacks \cite{choi2019evaluating,yin2018deep,choi2021deep,yan2022towards}, backdoor attacks against I2I networks have been left unstudied. This work conducts a comprehensive investigation of the vulnerability of I2I networks to backdoor attacks. As depicted in Figure \ref{Fig:backdoored image-to-image network}, we first introduce a backdoor attack targeting I2I networks. The backdoored I2I network functions normally when processing clean input images, i.e., yielding denoised or high-resolution images. However, it consistently exhibits backdoor behavior when the backdoor trigger appears in the input image, e.g., producing a specific inappropriate image. In addition, we further extend our I2I backdoor to attack downstream tasks (such as image classification and object detection), where the attacker has no knowledge of the downstream classifier or detector. As illustrated in Figure \ref{Fig:backdoored image-to-image network and downstream classification}, the upstream denoising network appears to function normally on input noisy images. However, the denoised version of the backdoor-triggered input image will induce a misclassification/misdetection\footnote{In this work, the target of the misdetection is to fabricate additional wrong detections (i.e., adding false positives).} of the clean classifier/detector with a high probability.

Compared with existing adversarial attacks against I2I networks that aim to degrade the quality of the output image \cite{yin2018deep,choi2019evaluating,choi2021deep,yan2022towards}, our proposed I2I backdoor attacks can lead to more severe consequences, e.g., outputting a backdoor target image that is completely irrelevant to the input image. It should be pointed out that the backdoor behavior of our I2I backdoor can also be configured to output lower quality images, such as increasing noise for the image denoising task, or outputting low-resolution images for the image super-resolution task. In contrast, outputting the backdoor target image is more challenging, it can lead to more serious security consequences and can also be used for some positive applications (see Figure \ref{Fig:I2I backdoor for image steganography}).

However, achieving such an I2I backdoor attack is non-trivial. Unlike backdoor attacks on classification models that map a triggered image to a specific target class, the mapping relationship in our I2I backdoor (i.e., the transformation of a triggered image to a specific backdoor target image) is notably more complicated. Directly using existing backdoor triggers for image classification tasks\footnote{We also employ existing backdoor triggers for image classification tasks to perform our I2I backdoor attack, refer to Section \ref{sec:Impact of the Backdoor Trigger} for detailed experimental results.} can not strike a good balance between preserving normal-functionality and enhancing attack effectiveness. Hence, we design a universal adversarial perturbation (UAP) generation algorithm for I2I networks, where the generated UAP is employed as the backdoor trigger for the I2I backdoor. After that, we employ a multi-task learning (MTL) framework, augmented with dynamic weighting methods, to accelerate the backdoor training process. In terms of the I2I backdoor attack against downstream tasks, we first introduce the UAP generation algorithm for downstream classification/detection models. Then we attach the UAP to the noise-free image and use this image as the backdoor target image to train the upstream backdoor image denoising model. Consequently, the denoised result of the backdoor-triggered image will contain the classification/detection UAP, thereby provoking misclassification or misdetection. 

Notably, this work focuses on I2I networks used for I2I tasks (such as image denoising and super-resolution) rather than image generative networks such as generative adversarial net (GAN) and diffusion model. There have been some works that explore the backdoor attacks on GAN \cite{salem2020baaan,rawat2022devil,jin2022backdoor} and diffusion model \cite{chen2023trojdiff,chou2023backdoor,struppek2022rickrolling}. However, backdoor attacks against GANs focused on modifying the loss functions of the generator and discriminator; backdoor attacks against diffusion models focused on manipulating the diffusion process. These backdoor methods cannot be applied and compared in our I2I backdoor attack, because I2I networks do not have generators or discriminators and do not involve a diffusion process.

In summary, our contributions are as follows:

\begin{itemize}[leftmargin=*,itemsep=0pt,topsep=0pt]
    \item We present the first backdoor attack against I2I networks. Specifically, to achieve a good balance between normal-functionality and attack effectiveness, we design a UAP generation algorithm for I2I networks and employ the generated UAP as the backdoor trigger for the I2I backdoor. After that, we employ multi-task learning (MTL) with dynamic weighting methods in the backdoor training process to obtain faster convergence rates.
    \item We further propose an I2I backdoor attack that is targeted at downstream tasks, including image classification and object detection. By utilizing the universal adversarial example against classification/detection models as the backdoor target image, the denoised result of the triggered image will induce a misclassification/misdetection with a high probability.
    \item We conduct extensive experiments on various state-of-the-art (SOTA) I2I architectures. The results demonstrate the effectiveness of our I2I backdoor attack against I2I tasks and downstream tasks. Besides, our approach exhibits remarkable robustness against diverse backdoor defenses.
\end{itemize}

The remainder of this paper is organized as follows: the background of this work is presented in Section \ref{sec:Background}. The threat model is described in Section \ref{sec:Threat Model}. Section \ref{sec:I2I Backdoor Attack against I2I Tasks} and \ref{sec:I2I Backdoor Attack that is Targeted at Downstream Classification Tasks} provide the details of our attack methodologies. Experimental evaluations are shown in Section \ref{sec:Evaluation Results}. Finally, Section \ref{sec:Conclusions} concludes the paper.

\section{Background}
\label{sec:Background}

\subsection{Image-to-image Networks}
\label{sec:Image-to-image Networks}

Owing to the remarkable advancements in deep learning within the field of computer vision, numerous deep learning-based Image-to-Image (I2I) network architectures have emerged to deal with a diverse range of I2I tasks, encompassing image super-resolution, image denoising, etc. For instance, \textit{Wang et al.} proposed ESRGAN \cite{Wang_2018_ECCV_Workshops}, which leverages relativistic GANs to enhance image super-resolution. DPIR, as proposed by  \textit{Zhang et al.} \cite{zhang2021plug}, offers a plug-and-play solution for image super-resolution, streamlining the super-resolution process. \textit{Zamir et al.} proposed MPRNet \cite{Zamir_2021_CVPR}, a multi-stage I2I architecture used for image restoration. \textit{Zhang et al.} proposed SCUNet \cite{zhang2022practical}, which combines the strengths of residual convolutional layers and Swin Transformer blocks \cite{liu2022swin}, yielding superior image denoising results. \textit{Zamir et al.} \cite{zamir2020learning} proposed MIRNet, which excels in feature extraction across multiple spatial scales, producing high-quality and high-resolution images.

In this work, we conduct comprehensive evaluations on these SOTA I2I architecture to investigate the vulnerability of I2I networks to backdoor attacks.


\subsection{Adversarial Attacks against I2I Networks}
\label{sec:Attacks against Image-to-image Networks}

Few works have delved into the susceptibility of I2I networks to adversarial attacks. For example, \textit{Yin et al.} \cite{yin2018deep} employed the gradient-based adversarial attacks in classification problems to attack the denoising networks with three downstream tasks: image style transfer \cite{gatys2016image}, image classification and image caption \cite{xu2015show}; \textit{Choi et al.} \cite{choi2019evaluating,choi2021deep} investigated adversarial attacks against various deep I2I networks including colorization networks, super-resolution networks, denoising networks and deblurring networks; \textit{Yan et al.} \cite{yan2022towards} proposed an adversarial attack against image denoising networks and developed an adversarial training strategy to enhance the robustness of denoising networks.

However, none of the existing studies explores backdoor attacks against I2I networks. In contrast to adversarial attacks, the I2I backdoor attacks proposed in this work exhibit more severe security threats and can be used for positive applications (see Figure \ref{Fig:I2I backdoor for image steganography}). This underscores the imperative need to investigate the vulnerability of I2I networks against backdoor attacks.


\subsection{Backdoor Attacks against Image Generative Networks}
\label{sec:Backdoor Attacks against Image Generative Networks}
Several works have explored backdoor attacks on generative models such as GAN \cite{salem2020baaan,rawat2022devil,jin2022backdoor} and diffusion model \cite{chen2023trojdiff,chou2023backdoor,struppek2022rickrolling}. Concretely, \textit{Salem et al.} \cite{salem2020baaan} and \textit{Rawat et al.} \cite{rawat2022devil} proposed backdoor attacks against GANs, where they modified the loss functions of the generator and discriminator to make GAN output the backdoor target image for the triggered input image; \textit{Jin et al.} \cite{jin2022backdoor} extended this backdoor attack in federated learning GAN; \textit{Chou et al.} \cite{chou2023backdoor} and \textit{Chen et al.} \cite{chen2023trojdiff} embedded backdoor in diffusion models by manipulating the diffusion process. 

Nevertheless, these backdoor methods cannot be applied and compared in our I2I backdoor attack, because I2I networks do not have a generator and discriminator and do not entail a diffusion process.


\section{Threat Model}
\label{sec:Threat Model}
In this work, we consider a malicious I2I network provider, who has control of the training process of the victim I2I network. The adversary trains the backdoored I2I network and makes it accessible for users to download. For the I2I backdoor attack that is targeted at downstream image classification and object detection tasks, the attacker has no knowledge of the downstream classifier or detector. The I2I backdoor attack must satisfy the following requirements:

\begin{itemize}[leftmargin=*,itemsep=0pt,topsep=0pt]
  \item \textbf{Normal-functionality}. The I2I backdoor must preserve the performance of the I2I network when processing clean input images. In the context of the I2I backdoor attack against I2I tasks, this requirement implies that the backdoored denoising/super-resolution network should output normal denoised/high-resolution images for clean input images. In the case of the I2I backdoor attack targeting downstream tasks, for clean input images and the backdoored upstream denoising model, the downstream classification/detection accuracy should be similar to that with the clean upstream denoising model.
  \item \textbf{Effectiveness.} For the I2I backdoor attack against I2I tasks, the backdoored I2I model should be capable of generating the backdoor target image when processing images with the backdoor trigger. In the context of the I2I backdoor attack against downstream classification/detection tasks, for backdoor-triggered input images and the backdoored upstream denoising model, the denoised images should provoke misclassification/misdetection by the downstream classification/detection model with a high probability.

\end{itemize}


\section{I2I Backdoor Attack against I2I Tasks}
\label{sec:I2I Backdoor Attack against I2I Tasks}
\begin{figure*}
\centering
\includegraphics[width=0.9\textwidth]{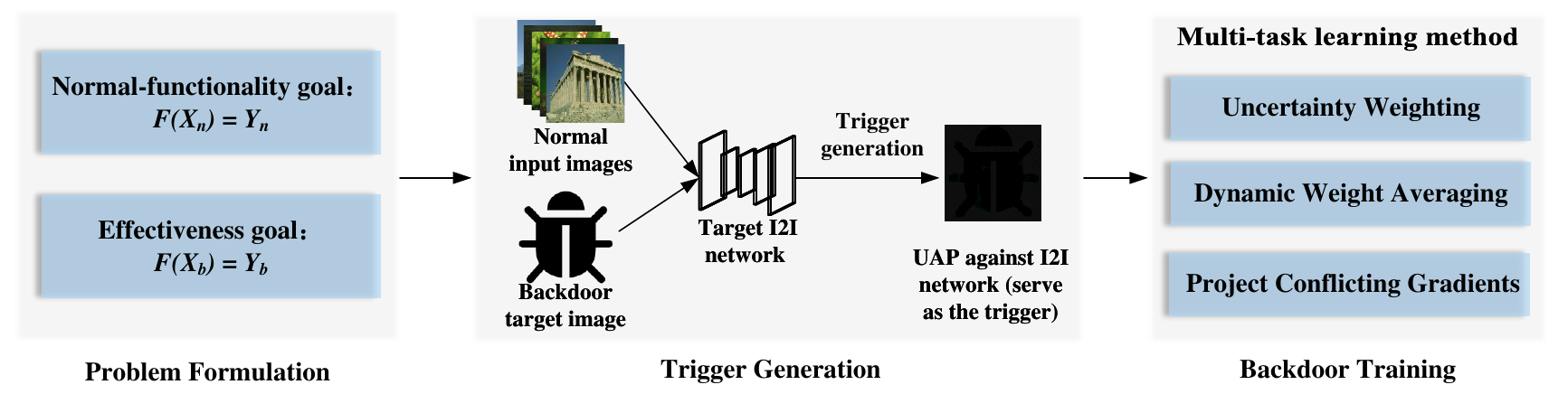}
\caption{The workflow of I2I backdoor attack.}
\label{Fig:The workflow of backdoor attack against I2I tasks}
\end{figure*}

In this section, we present the details of I2I backdoor attack against I2I tasks. The workflow is illustrated in Figure \ref{Fig:The workflow of backdoor attack against I2I tasks}.

For ease of reference, we also provide the notations used in this work in TABLE \ref{Notations and description}.
\begin{table}[H]\small
\centering
\caption{Notations and description}
\label{Notations and description}
\begin{tabular}{cc}
  \toprule
  Notations &  Description \\
  \hline
  $X_n$ &    the normal input image \\
  $X_b$ &    the triggered input image \\
  $Y_n$ &    the normal target image \\
  $Y_b$ &  the backdoor target image \\
  $t$ &    trigger for the I2I task \\
  $F$ &  the targeted I2I model \\
  $S$ &  a small set of normal images \\
  $s$ &  update step size of the trigger \\
  $I$ & the maximal number of iterations \\
  $\mathcal{L}_{t}$ &  loss function to optimize the trigger \\
  $\mathcal{L}_{m}$ &  loss function of the main task \\
  $\mathcal{L}_{b}$ &  loss function of the backdoor task \\
  $C$ &   the downstream image classification model \\
  $u$ &   trigger for the image classification task \\
  $D$ &   the downstream object detection model \\
  \bottomrule
\end{tabular}
\end{table}

\subsection{Problem Formulation}
\label{sec:Problem Formulation against I2I Tasks}
We denote $X_n$ as the normal input image (i.e., the normal low-resolution/noisy image), $Y_n$ as the normal output image (i.e., the high-resolution/noise-free image), $X_b$ as the backdoor-triggered input image, $Y_b$ as the backdoor target image\footnote{In this work, we choose a bug image as the backdoor target image, which is completely irrelevant to the input image.}, $F$ as the target I2I network. According to the requirements described in Section \ref{sec:Threat Model}, the goal of the I2I backdoor against I2I tasks can be formulated as:
\begin{equation}
\label{eq:normal-functionality goal of the I2I backdoor attack}
\text{Normal-functionality goal:} \; F(X_n) = Y_n \\
\end{equation}
\begin{equation}
\label{eq:Effectiveness goal of the I2I backdoor attack}
\text{Effectiveness goal:} \; F(X_b) = Y_b \\
\end{equation}


\subsection{Backdoor Trigger}
\label{sec:Backdoor Trigger against I2I Tasks}

In backdoor attacks against classification models, numerous studies \cite{garg2020can,zhong2020backdoor,lederer2023silent,phan2022invisible,zhao2020clean} have opted for the utilization of the targeted Universal Adversarial Perturbation (UAP) as the backdoor trigger. This is because the targeted UAP is able to push the classification results of triggered samples to the backdoor target class and is therefore more facilitating for backdoor embedding.

Inspired by this idea, we design a targeted UAP attack for I2I networks and use this UAP as the trigger for our I2I backdoor. As presented in Algorithm \ref{alg:Trigger Generation Algorithm}, for a small set of normal input images $S$, we iteratively pick one sample ($X_i$) from $S$ and employs the gradient descent algorithm to minimize $\mathcal{L}_{t}$ for $I$ rounds to optimize trigger $t$:

\begin{equation}
\label{eq:trigger loss}
\mathcal{L}_{t}=\|F(X_i+t)-Y_b\|_2
\end{equation}
The optimizing process is performed for all samples in $S$ one by one and the final $t$ is returned as the UAP trigger.

\begin{algorithm}
\caption{The Generation Algorithm of the UAP Trigger}
\label{alg:Trigger Generation Algorithm}
\begin{algorithmic}[1]
\REQUIRE a small set of normal input images $S$; the victim I2I model $F$; the update step size of the trigger $s$; maximal number of iterations $I$; the backdoor target image $Y_b$; the range of the trigger $(-\epsilon_t,+\epsilon_t)$.
\ENSURE the UAP trigger $t$
\STATE randomly initialize $t$;
\FOR{ each sample $X_i \in S $}
\STATE $j\leftarrow0$ (iteration counter)
\WHILE {$j<=I$}
\STATE $\Delta = \frac{\partial \mathcal{L}_{t}}{\partial t} $
\STATE   $t \leftarrow t-s*sign(\Delta), \ t \leftarrow clip(t,-\epsilon_t,+\epsilon_t)$
\STATE  $j\leftarrow j+1$
\STATE  Update $\mathcal{L}_{t}$ According to Eq.(\ref{eq:trigger loss})
\ENDWHILE
\ENDFOR
\STATE \textbf{return} $t$
\end{algorithmic}
\end{algorithm}

\begin{figure}
  \centering
  \subfigure[No trigger]{
    \includegraphics[width=0.22\linewidth]{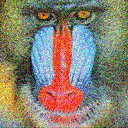}
    }
    \subfigure[Patch ]{
    \includegraphics[width=0.22\linewidth]{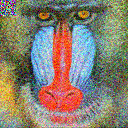}
    }
    \subfigure[Color ]{
    \includegraphics[width=0.22\linewidth]{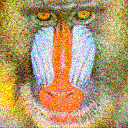}
    }
    \subfigure[Blend ]{
    \includegraphics[width=0.22\linewidth]{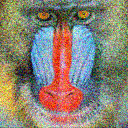}
    }
    \subfigure[Gaussian ]{
    \includegraphics[width=0.22\linewidth]{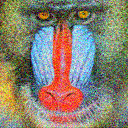}
    }
    \subfigure[Filter ]{
    \includegraphics[width=0.22\linewidth]{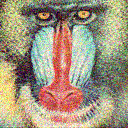}
    }
    \subfigure[Refool ]{
    \includegraphics[width=0.22\linewidth]{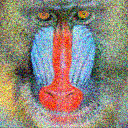}
    }
    \subfigure[UAP ]{
    \includegraphics[width=0.22\linewidth]{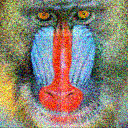}
    }
  \caption{Visual examples of input noisy images with/without trigger.}
  \label{Fig:Examples of triggered images}
\end{figure}

In our experiments, we also employ existing backdoor triggers for image classification tasks to perform our I2I backdoor attack, including patch trigger \cite{gu2019badnets}, blend trigger \cite{chen2017targeted}, refool trigger \cite{liu2020reflection}, color trigger \cite{jiang2023color}, Instagram filter trigger \cite{liu2019abs} and Gaussian noise trigger \cite{chen2021use}. Figure \ref{Fig:Examples of triggered images} illustrates the input noisy images with these backdoor triggers. The experimental results in Section \ref{sec:Impact of the Backdoor Trigger} demonstrate the superiority of our designed UAP trigger.

\subsection{Backdoor Training}
\label{sec:Backdoor Training against I2I Tasks}
\subsubsection{Backdoor Training with Multi-task Learning (MTL)}
After identifying the backdoor trigger pattern, the subsequent step is to embed the backdoor into the I2I model via the backdoor training process. In order to accomplish the dual objectives of ensuring normal-functionality and enhancing attack effectiveness simultaneously, we have devised two loss functions for the main task and the backdoor task. After that, we leverage the multi-task learning (MTL) framework to conduct the backdoor training process.

\textbf{The main task} is to satisfy the normal-functionality goal, i.e., the backdoored model is expected to perform normally on normal input images. The loss function can be defined as:
\begin{equation}
\label{eq:loss function of the main task}
\mathcal{L}_{m} = \| F(X_n) - Y_n \|_2\\
\end{equation}

\textbf{The backdoor task} is to achieve the attack effectiveness goal, i.e., the backdoored model is expected to output the backdoor target image for the backdoor-triggered input image. The loss function can be formulated as:

\begin{equation}
\label{eq:loss function of the backdoor task}
\mathcal{L}_{b} = \|F(X_b)- Y_b\|_2\\
\end{equation}

Therefore, the total loss for backdoor training can be formulated as:
\begin{equation}
\label{eq:total loss}
\mathcal{L}_{total}=\mathcal{L}_{m} + \mathcal{L}_{b}
\end{equation}

\subsubsection{Dynamic Weighting Methods}

However, in the training process with Equation (\ref{eq:total loss}), the $\mathcal{L}_{total}$ is prone to be dominated by the task with a larger loss and fall into the local optimum, resulting in lower attack performance. This is attributed to the complicated mapping relationship in the I2I backdoor attack (backdoor-triggered images to the backdoor target image), making it difficult to balance the two tasks. Hence, we employ SOTA weighting methods, including Uncertainty Weighting (UW) \cite{kendall2018multi}, Dynamic Weight Averaging (DWA) \cite{liu2019end} and Project Conflicting Gradients (PCGrad) \cite{yu2020gradient} in the MTL process to avoid local optimum and accelerate convergence rates. Below we describe how to employ these weighting methods in our backdoor training process.

\textbf{UW} assigns larger weights to ``easier'' tasks, where it employs homoscedastic task uncertainty to balance different loss functions of different tasks. The $L_{total}$ in this work can be formulated as:
\begin{equation}
\label{eq:Uncertainty Weighting}
\mathcal{L}_{total}=\frac{1}{2 \sigma_m^2} \mathcal{L}_{m}+\frac{1}{2 \sigma_b^2} \mathcal{L}_{b}+\log \sigma_m\sigma_b
\end{equation}
where $\sigma_m$ and $\sigma_b$ represent the variance of $\mathcal{L}_{m}$ and $\mathcal{L}_{b}$. For the task with large uncertainty (i.e., large variance), the corresponding weights of its loss function are correspondingly reduced. The function of $\log \sigma_{m,b}$ is to prevent $\sigma_{m,b}$ from being too large.

\textbf{DWA} forces each task to learn at a similar rate. The weight of each task is formulated as follows:
\begin{equation}
\label{eq:DWA1}
w_i(t)=\frac{N e^{\left(r_i(t-1) / T\right)}}{\sum_{n=1}^N e^{\left(r_n(t-1) / T\right)}}, r_i(t-1)=\frac{\mathcal{L}_i(t-1)}{\mathcal{L}_i(t-2)}
\end{equation}
where $w_i(t)$ represents the weight of task $i$ at step $t$, $N$ represents the total number of the tasks, $r_n(t)$ is the ratio of the current loss to the previous loss, $T$ is the temperature-scaling hyperparameter \cite{hinton2015distilling}, which controls the softness of task weighting.


\textbf{PCGrad} is designed to address the challenging issue of gradient conflict. Specifically, during our backdoor training process, it is common that the gradients of the main task and the backdoor task exhibit some degree of conflict\footnote{When the cosine similarity between the two gradients $\cos\theta<0$, the two gradients are considered to be conflicted.}. This conflict often results in sluggish convergence rates or diminished attack performance. For every training batch, PCGrad calculates the cosine similarity between the gradient of the main task, denoted as $\mathbf{g}_m$, and the gradient of the backdoor task, represented as $\mathbf{g}_b$. In cases where the gradients are not conflicted, they remain unaltered. When conflicts arise, PCGrad replaces $\mathbf{g}_b$ with its projection onto the normal plane of $\mathbf{g}_m$, as presented in Equation (\ref{eq:PCGrad}). This mechanism enhances the backdoor training process, mitigating gradient conflicts and fostering more efficient convergence and heightened attack performance.

\begin{equation}
\label{eq:PCGrad}
\mathbf{g}_b=\mathbf{g}_b-\frac{\mathbf{g}_b \cdot \mathbf{g}_m}{\|\mathbf{g}_m\|^2} \mathbf{g}_m
\end{equation}

In Section \ref{sec:Evaluation Results}, we conduct extensive ablation studies to rigorously assess the performance of these dynamic weighting methods.


\section{I2I Backdoor Attack that is Targeted at the Downstream Tasks}
\label{sec:I2I Backdoor Attack that is Targeted at Downstream Classification Tasks}

In addition to attacking I2I tasks, we further introduce an I2I backdoor attack that is targeted at the downstream image classification or object detection tasks. Specifically, we first embed the backdoor into the upstream image denoising model. Consequently, the denoised version of the backdoor-triggered image will be misclassified/misdetected by normal downstream classification/detection models with a high probability.

\subsection{Targeted at Downstream Image Classification Task}
According to the requirements described in Section \ref{sec:Threat Model}, the normal-functionality and effectiveness goal of the I2I backdoor attack against downstream classification task can be formulated as:
\begin{equation}
\label{eq:normal-functionality goal of the I2I backdoor attack against downstream classification}
\text{Normal-functionality goal:} \; C(F(X_n)) = C(Y_n)\\
\end{equation}
\begin{equation}
\label{eq:Effectiveness goal of the I2I backdoor attack against downstream classification}
\text{Effectiveness goal:} \; C(F(X_b)) \neq C(Y_n)
\end{equation}
where $C$ is the normal downstream image classification model.



For backdoor trigger types and backdoor training methods, we adopt the same attack configurations described in Section \ref{sec:Backdoor Trigger against I2I Tasks} and \ref{sec:Backdoor Training against I2I Tasks}. Differently, for the input backdoor-triggered image, we attach the classification UAP\footnote{The classification UAP is designed to induce misclassifications of classification models, which is different from the UAP against I2I networks in Section \ref{sec:Backdoor Trigger against I2I Tasks}.} to its noise-free version and use this image as the backdoor target image. Consequently, the denoised version of the input backdoor-triggered image will contain the classification UAP thereby leading to a misclassification.

Specifically, the generation algorithm of the classification UAP is presented in Algorithm \ref{alg:The Generation Algorithm of UAP against classification models}. For each sample ($X_i$) in the dataset $S$, the algorithm first determines whether $X_i+u$ is able to cause the misclassification of the model $C$. If not, the algorithm performs an adversarial attack algorithm (such as DeepFool \cite{moosavi2016deepfool}, PGD \cite{madry2018towards}) to optimize the $u$ so that $X_i+u$ crosses the classification boundary. The optimizing process is conducted for all samples in $S$ and the final $u$ is returned as the classification UAP.

\begin{algorithm}
\caption{The Generation Algorithm of the Classification UAP}
\label{alg:The Generation Algorithm of UAP against classification models}
\begin{algorithmic}[1]
\REQUIRE a small set of normal input images $S$; a clean pre-trained downstream classification model $C$; the range of the classification UAP $(-\epsilon_u,+\epsilon_u)$.
\ENSURE the UAP for image classification $u$
\STATE initialize $u \leftarrow 0$;
\FOR{ each sample $X_i \in S $}
\IF {$C(X_i+u)=C(X_i)$}
\STATE Compute the minimal perturbation that sends $X_i+u$ to the decision boundary: \\ $\Delta u_i=\arg \min _r\|r\|_2, \ \text{s.t.} \  C\left(X_i+u+r\right)\neq C(X_i)$
\STATE Update the perturbation: \\ $u \leftarrow u+\Delta u_i, \ u \leftarrow clip(u,-\epsilon_u,+\epsilon_u)$
\ENDIF
\ENDFOR
\STATE \textbf{return} $u$
\end{algorithmic}
\end{algorithm}

\subsection{Targeted at Downstream Object Detection Task}
Similarly, the normal-functionality and effectiveness goal of the I2I backdoor attack against downstream detection task can be defined as Equation (\ref{eq:normal-functionality goal of the I2I backdoor attack against downstream detection}) and (\ref{eq:Effectiveness goal of the I2I backdoor attack against downstream detection}), respectively.
\begin{equation}
\label{eq:normal-functionality goal of the I2I backdoor attack against downstream detection}
\text{Normal-functionality goal:} \; D(F(X_n)) = D(Y_n)
\end{equation}
\begin{equation}
\label{eq:Effectiveness goal of the I2I backdoor attack against downstream detection}
\text{Effectiveness goal:} \; D(F(X_b)) \neq D(Y_n)
\end{equation}
where $D$ is the normal downstream object detection model.


We also employ the same backdoor trigger types and backdoor training methods described in Section \ref{sec:Backdoor Trigger against I2I Tasks} and \ref{sec:Backdoor Training against I2I Tasks}. For the backdoor target image, we first adopt the existing universal adversarial attack against object detection \cite{chow2020adversarial} to generate the UAP\footnote{The detection UAP is designed to fabricate additional wrong detections (i.e., adding false positives).} for the object detection task. After that, for the backdoor-triggered image, we attach the detection UAP to its noise-free image and use this image as the backdoor target image.

It is worth noting that the attacker does not need to have knowledge of the downstream classification/detection model. It can utilize surrogate models to execute the UAP generation algorithm. Due to the transferability of UAP, images with the UAP are able to induce misclassifications/misdetections of other clean classification/detection models. 


\section{Evaluation}
\label{sec:Evaluation Results}
We perform extensive experiments over different datasets and I2I networks to evaluate the performance of our I2I backdoor attacks. All experiments are implemented in Python and run on a NVIDIA RTX A6000. 


\subsection{Experimental Setup}
\label{sec:Experimental Setup}

\subsubsection{Model Architecture}
\begin{itemize}[leftmargin=*,itemsep=0pt,topsep=0pt]
\item \textbf{I2I backdoor against I2I tasks:} this work considers the two most commonly used I2I tasks (image denoising and image super-resolution) as examples to investigate the vulnerability of I2I networks to backdoor attacks. For the two tasks, we have selected several state-of-the-art (SOTA) I2I network architectures, including SCUNet \cite{zhang2022practical}, MPRNet \cite{Zamir_2021_CVPR}, MIRNet \cite{zamir2020learning}, DPIR \cite{zhang2021plug} and ESRGAN \cite{Wang_2018_ECCV_Workshops}, for experimental evaluations. We firmly believe that other I2I tasks and I2I network architectures are also susceptible to the I2I backdoor attacks in this work. 
\item \textbf{I2I backdoor against downstream tasks:} in the context of the I2I backdoor that targets at the downstream classification/detection task, we employ the aforementioned image denoising networks to conduct the upstream image denoising task. For the downstream classification task, we use the pre-trained ResNet50, VGG19 and MobileNetv2 model to perform image classification; for the downstream detection task, we use the pre-trained MobileNet-YOLOv3, EfficientNet-YOLOv3 and Darknet53-YOLOv3 model to perform object detection.
\end{itemize}

\subsubsection{Datasets}
\begin{itemize}[leftmargin=*,itemsep=0pt,topsep=0pt]
\item \textbf{Image denoising task:} following previous works \cite{zhang2017beyond,chen2016trainable,schmidt2014shrinkage}, we use Color400 as the training data, and CSet8 as the testing data. 
\item \textbf{Image super-resolution task:} we choose BSD100 \cite{martin2001database} as the training data, and Set14 \cite{zeyde2012single} as the testing data.
\item \textbf{Downstream image classification task:} we evaluate our I2I backdoor against the downstream image classification task on the ImageNet-1k \cite{ILSVRC15} dataset.
\item \textbf{Downstream object detection task:} we evaluate our I2I backdoor against the downstream object detection task on the Pascal VOC dataset \cite{everingham2015pascal}. 
\end{itemize}

\subsubsection{Attack Configuration}
\begin{itemize}[leftmargin=*,itemsep=0pt,topsep=0pt]
\item \textbf{Baseline trigger settings:} (1) Patch backdoor trigger: following previous work \cite{gu2019badnets}, we employ a 16*16 patch in the top left corner of the input 128*128 image as the backdoor trigger; (2) Gaussian backdoor trigger: following previous work \cite{chen2021use}, we employ a specific Gaussian noise as the backdoor trigger. The mean is set to 0, the standard deviation is set to 20/255; (3) Color backdoor trigger: we follow the attack configuration proposed in \cite{jiang2023color} and use a specific shift in the color space as the backdoor trigger; (4) Blend backdoor trigger: we follow the attack configuration proposed in \cite{chen2017targeted} and use the backdoor target image as the blend image. The blending ratio is set to 0.1; (5) Refool backdoor trigger: we follow the attack configuration proposed in \cite{liu2020reflection} and use the backdoor target image as the reflection trigger. The reflection ratio is set to 0.1; (6) Filter backdoor trigger: we follow the attack configuration proposed in \cite{liu2019abs} and use a specific Instagram filter as the backdoor trigger.
\item \textbf{UAP trigger settings:} the number of normal images in $D$ is set to 10, the update step size of the trigger $s$ is set to 5/255, the maximum number of iterations $I$ is set to 20, the range of the trigger is set to (-15/255, +15/255). 
\item \textbf{Backdoor training process settings:} we follow the hyperparameter settings in UW \cite{kendall2018multi}, DWA \cite{liu2019end} and PCGrad \cite{yu2020gradient}, and train the backdoor model with the Adam optimizer (learning rate equals to 0.0001). Besides, we also introduce a static weighting (SW) approach for backdoor training as a baseline comparison, where the weight of the main task is set to 0.9 and the weight of the backdoor task is set to 0.1. 
\end{itemize}

\subsubsection{Evaluation Metrics}
\begin{itemize}[leftmargin=*,itemsep=0pt,topsep=0pt]
\item \textbf{I2I backdoor against I2I tasks:} we employ the SSIM \cite{wang2004image} between images to measure the attack performance. Specifically, for the backdoored I2I model, we calculate the SSIM between the denoised/super-resolved result for clean input image and the ground truth image (i.e., noise-free image or high-resolution image) to evaluate the normal-functionality; we calculate the SSIM between the denoised/super-resolved result for triggered image and the backdoor target image to evaluate the attack effectiveness.
\item \textbf{I2I backdoor against downstream tasks:} for the backdoored upstream image denoising model, we calculate the test accuracy and mean Average Precision (mAP) of the downstream classification and detection model on the denoised results for normal input images to measure the normal-functionality, respectively; we calculate the attack success rate (ASR) of the denoised results for triggered images on the downstream model to evaluate the attack effectiveness.

\item \textbf{Stealthiness of triggered images:} to evaluate the stealthiness of triggered images, we employ LPIPS \cite{zhang2018unreasonable} distance as the metric, which is more consistent with human vision system. Concretely, LPIPS evaluates the perceptual distance between two images by means of a deep learning model. LPIPS suggests that even if two images are very close to each other at the pixel level, a human observer may perceive them as different. Therefore, LPIPS uses pre-trained deep networks (e.g., VGG, AlexNet) to extract image features\footnote{In this work, we employ the pre-trained VGG network for experimental evaluations.}, and then calculates the distance between these features to evaluate the perceptual similarity between images. It is formulated as Equation (\ref{eq:LPIPS}):
\begin{equation}\small
\label{eq:LPIPS}
lpips(x, \hat{x}) = \sum_{l} \frac{1}{H_l W_l} \sum_{h=1}^{H_l} \sum_{w=1}^{W_l} \left\| \mathbf{w}_l \odot \left( \hat{\mathbf{f}}_l^x(h,w) - \hat{\mathbf{f}}_l^{\hat{x}}(h,w) \right) \right\|_2^2
\end{equation}
where $\hat{\mathbf{f}}_l^x(h,w)$ represents the features extracted from image $x$ at layer $l$; $\mathbf{w}_l$ represents the channel weight of layer $l$; $H_l, W_l$ represent the height and width of the feature map in layer $l$; $\odot$ denotes the channel-by-channel multiplication and $\| \cdot \|_2^2$ denotes the square of the Euclidean distance. A smaller LPIPS value means that the two images are more similar in terms of human visual perception. 
\end{itemize}


\subsection{Attack Performance Evaluation}
\subsubsection{Main Results}
\label{sec:Impact of the Backdoor Trigger}

\begin{table*}\small
\centering
\caption{The performance of I2I backdoor with different triggers and MTL methods on image denoising task.} 
\label{Table: Performance of I2I backdoor with different backdoor triggers and MTL methods on image denoising task.}
\resizebox{0.77\linewidth}{!}{
\begin{threeparttable}
\begin{tabular}{ccc|cccccccc}
\toprule
 \multirow{2}{*}{Architecture} &  MTL  &   \multirow{2}{*}{SSIM\tnote{*}}  &  \multicolumn{8}{c}{Trigger type}  \\ \cline{4-11}
   &   method &     & None &Gaussian & Color & Filter & Patch & Blend & Refool & UAP\\
\midrule
 \multirow{12}{*}{DPIR} 
& \multirow{3}{*}{SW}  & Normal.   & 0.8936  &  0.8690   & 0.8872  &  0.7816  &0.7440  &0.8914   &0.8948  &  0.8961 \\
&   & Effect.  &  $\backslash$ &   0.9949 &   0.9964 &  0.8723 &  0.9992& 0.9437  &0.9938  & 0.9936\\ 
&   & Total  &  $\backslash$ &1.8649   & 1.8836   & 1.6539  & 1.7432  &1.8351  &1.8886  &  \textbf{1.8897} \\ \cline{2-11}
 & \multirow{3}{*}{DWA} & Normal.   & $\backslash$  &0.8660   &   0.7062 &  0.6351  &  0.7110  & 0.8766  & 0.8170  &  0.8855 \\
&   & Effect.  &  $\backslash$ & 0.9939  & 0.9786  &  0.8675 &  0.9995 & 0.9925 & 0.9158  & 0.9841\\ 
&   & Total  &  $\backslash$ & 1.8599   &  1.6848  &1.5026    & 1.7105 & 1.8691 & 1.7328 & \textbf{1.8696}\\ \cline{2-11}
 &  \multirow{3}{*}{UW} & Normal.   &$\backslash$  &  0.8661  &  0.8288  & 0.5853  & 0.6957 & 0.8821 &  0.8821  &  0.8850 \\
&   & Effect.  &  $\backslash$ & 0.9974  &  0.9928  &0.9799   & 0.9989  & 0.9767 & 0.9350 & 0.9898\\ 
&   & Total  &  $\backslash$ &1.8635   & 1.8216  &1.5652   &  1.6946 &   1.8588&1.8171  &  \textbf{1.8748} \\ \cline{2-11}
&  \multirow{3}{*}{PCGrad} & Normal.   &$\backslash$  &0.8649    & 0.7953   &  0.6800  &0.7372  & 0.8832  & 0.8863   & 0.8939  \\
&   & Effect.  &  $\backslash$ &  0.9970 &  0.9900 &  0.9839 & 0.9996 & 0.9302 &  0.9851&  0.9991\\ 
&   & Total  &  $\backslash$ & 1.8619  &1.7853   &  1.6639  & 1.7368  & 1.8134 &   1.8714& \textbf{1.8930}\\ \cline{1-11}

\multirow{12}{*}{SCUNet} 
& \multirow{3}{*}{SW}  & Normal.   & 0.8839  &0.8688   &0.7713   &0.7561   &0.7534   & 0.8827  &  0.8746  &  0.8834 \\
&   & Effect.  &  $\backslash$ &  0.9925 & 0.9906  & 0.8977  &   0.9414& 0.9499 &0.8268  &  0.9872\\ 
&   & Total  &  $\backslash$ &1.8613    &1.7619   &  1.6538  &1.6948  &1.8326   &1.7014  & \textbf{1.8705}\\ \cline{2-11}
 & \multirow{3}{*}{DWA} & Normal.  & $\backslash$  &0.8778   & 0.7360   & 0.8346 & 0.8228 & 0.8492 &0.7948  &  0.8803 \\
&   & Effect.  &  $\backslash$ & 0.9988  & 0.9901  &0.9725   &0.9927  &  0.8842& 0.8529 & 0.9988\\ 
&   & Total  &  $\backslash$ &1.8766   &1.7261   &   1.8071  & 1.8155  &1.7334   & 1.6477  & \textbf{1.8791} \\ \cline{2-11}
 &  \multirow{3}{*}{UW} &Normal.  &$\backslash$  & 0.8623   & 0.7536  & 0.7269  & 0.8087  & 0.8727   & 0.8612  &   0.8750 \\
&   & Effect.  &  $\backslash$ &  0.9988 &0.9996   &   0.9797 & 0.9965 &0.9745  & 0.9848 & 0.9980 \\ 
&   & Total  &  $\backslash$ & 1.8611  &  1.7532  &1.7066    & 1.8052 & 1.8472 &1.8460  & \textbf{1.8730}  \\ \cline{2-11}
&  \multirow{3}{*}{PCGrad} & Normal.   &$\backslash$  &  0.8752  &0.7511  &0.6722   &0.7667   & 0.8759  & 0.8371  & 0.8753 \\
&   & Effect.  &  $\backslash$ & 0.9990  & 0.9994  &   0.9849 & 0.9910  &0.9093  & 0.9230 &0.9986 \\ 
&   & Total  &  $\backslash$ & \textbf{1.8742}  &1.7505     & 1.6571   & 1.7577 &1.7852  &   1.7601&  1.8739\\ \cline{1-11}

\multirow{12}{*}{MPRNet} 
& \multirow{3}{*}{SW}  & Normal.  & 0.9081  & 0.7066   &  0.8865  &  0.7073 &0.7822   & 0.8638  &0.8938   &  0.9008 \\
&   & Effect.  &  $\backslash$ & 0.9704  &  0.8790 & 0.8729  & 0.7385  &0.9894  &0.9884  & 0.9977\\ 
&   & Total  &  $\backslash$ &1.6770   & 1.7655  & 1.5802   &  1.5207& 1.8532 &1.8822  & \textbf{1.8985}\\ \cline{2-11}
 & \multirow{3}{*}{DWA} & Normal.  & $\backslash$  & 0.6337   &  0.7701  & 0.6546   &0.7410  & 0.8890 & 0.8145  &  0.8721 \\
&   & Effect.  &  $\backslash$ &  0.9361 &  0.9322 & 0.9022  & 0.9676 &0.9970  & 0.9926 &0.9871 \\ 
&   & Total  &  $\backslash$ &   1.5698&1.7023   &  1.5568 &1.7086   &1.8860  & 1.8071 & \textbf{1.8592}\\ \cline{2-11}
 &  \multirow{3}{*}{UW} & Normal.  &$\backslash$  & 0.7354   &  0.7594  & 0.7256  &  0.7374  & 0.8867  & 0.8829   &  0.9079  \\
&   & Effect.  &  $\backslash$ &  0.9978 &   0.8723 &   0.9271 & 0.9943 & 0.9953 & 0.9964 & 0.9997\\ 
&   & Total  &  $\backslash$ & 1.7332  &  1.6317 & 1.6527  & 1.7317 & 1.8820 & 1.8793 & \textbf{1.9076}\\ \cline{2-11}
&  \multirow{3}{*}{PCGrad} & Normal.  &$\backslash$  &  0.7240  &  0.7430  &  0.7390&0.7452   &0.8853  & 0.8831 &   0.9151 \\
&   & Effect.  &  $\backslash$ & 0.9963  & 0.8587  & 0.9305  & 0.9884 & 0.9944 & 0.9957 & 0.9995 \\ 
&   & Total  &  $\backslash$ & 1.7203  &  1.6017 &   1.6695  &1.7336  & 1.8797 & 1.8788  &\textbf{1.9146} \\ \cline{1-11}

\multirow{12}{*}{MIRNet} 
& \multirow{3}{*}{SW}  & Normal.   &  0.9172 &  0.6887  &  0.8163  & 0.7797  &0.8546   & 0.8957  &0.8951  &0.9139   \\
&   & Effect.  &  $\backslash$ & 0.9983  & 0.9975  &   0.9534 & 0.9727 & 0.9779 & 0.9841 &0.9964 \\ 
&   & Total  &  $\backslash$ & 1.6870  & 1.8138  & 1.7331   &  1.8273& 1.8736 &  1.8792  & \textbf{1.9102}\\ \cline{2-11}
 & \multirow{3}{*}{DWA} & Normal.   & $\backslash$  & 0.7025  &  0.7464  & 0.8010  & 0.8147 & 0.8382  & 0.8220  &  0.8645 \\
&   & Effect.  &  $\backslash$ & 0.9921  &  0.9516 & 0.9033  & 0.9892 &  0.9497& 0.9845 &  0.9939\\ 
&   & Total  &  $\backslash$ & 1.6946   & 1.6980  &  1.7043  & 1.8039  &  1.7879& 1.8066 & \textbf{1.8585}\\ \cline{2-11}
 &  \multirow{3}{*}{UW} &Normal.   &$\backslash$  &  0.7205 &  0.8201 &  0.8101 & 0.8496 & 0.8872  &  0.8839 &  0.9001   \\
&   & Effect.  &  $\backslash$ &0.9811   &  0.9920 & 0.9720  & 0.9956 & 0.9905 & 0.9657 & 0.9920\\ 
&   & Total  &  $\backslash$ &  1.7016  &  1.8121 &  1.7821  &1.8452  & 1.8777 & 1.8496  & \textbf{1.8921}\\ \cline{2-11}
&  \multirow{3}{*}{PCGrad} & Normal.   &$\backslash$  &0.6991   &    0.8363 & 0.8325 & 0.8575  &  0.8944 & 0.8997  &  0.9060  \\
&   & Effect.  &  $\backslash$ &  0.9954 &  0.9719 &0.9819   &0.9939  & 0.9823 &0.9660  & 0.9994\\ 
&   & Total  &  $\backslash$ &  1.6945  &  1.8082 & 1.8144   & 1.8514 & 1.8767  &  1.8675 & \textbf{1.9054}\\  \cline{1-11}

\multirow{12}{*}{ESRGAN} 
& \multirow{3}{*}{SW}  & Normal.  &  0.9112 & 0.6009  & 0.7739   & 0.8056   & 0.6497 &0.8726   & 0.8667 &0.9146   \\
&   & Effect.  &  $\backslash$ & 0.9733  & 0.9713  &  0.9345 & 0.9969 & 0.8121 &0.9330  &0.9925\\ 
&   & Total  &  $\backslash$ &   1.5742 &1.7452   &  1.7401 & 1.6466  & 1.6847 & 1.7997  & \textbf{1.9071}\\ \cline{2-11}
 & \multirow{3}{*}{DWA} & Normal.   & $\backslash$  &  0.5565  & 0.7205 &0.6585    &  0.5707  &0.7926 &  0.8518 & 0.9073\\
&   & Effect.  &  $\backslash$ &0.9745   & 0.9903  &   0.7642 &0.9569  & 0.7870 &  0.8895 &  0.9470\\ 
&   & Total  &  $\backslash$ & 1.5310  & 1.7108   &  1.4227 & 1.5276 &  1.5796 &  1.7413 & \textbf{1.8544}\\ \cline{2-11}
 &  \multirow{3}{*}{UW} & Normal.   &$\backslash$  & 0.6037  & 0.7944  &  0.6443  & 0.6220 & 0.8579 & 0.8715 &  0.8962   \\
&   & Effect.  &  $\backslash$ &0.9956   &   0.9986 &  0.9772 &0.9886  & 0.9807 &0.9881  & 0.9985\\ 
&   & Total  &  $\backslash$ & 1.5993  & 1.7930  &  1.6215 &1.6106   &  1.8386 &  1.8596  & \textbf{1.8948}\\ \cline{2-11}
&  \multirow{3}{*}{PCGrad} & Normal.  &$\backslash$  & 0.5912   &  0.7892 &0.7715   &  0.4988  &0.8679   &  0.8807   & 0.9086  \\
&   & Effect.  &  $\backslash$ &  0.9968 & 0.9990  & 0.9872  & 0.9729 &0.9767  &  0.9891& 0.9992\\ 
&   & Total  &  $\backslash$ &  1.5880 & 1.7882   &   1.7587 &1.4717  &  1.8446& 1.8698 & \textbf{1.9078}\\

\bottomrule
\end{tabular}
 \begin{tablenotes}
        \footnotesize
        \item[*] Normal. denotes the normal-functionality; Effect. denotes the effectiveness; Total represents the sum of them. The bolded results represent the maximum total score.
      \end{tablenotes}
    \end{threeparttable}}
\vspace{-10pt}
\end{table*}

\begin{table*}\small
\centering
\caption{The performance of I2I backdoor with different triggers and MTL methods on image super-resolution task.} 
\label{Table: Performance of I2I backdoor with different backdoor triggers and MTL methods on image super-resolution task.}
\resizebox{0.77\linewidth}{!}{
\begin{threeparttable}
\begin{tabular}{ccc|cccccccc}
\toprule
 \multirow{2}{*}{Architecture} &  MTL  &   \multirow{2}{*}{SSIM\tnote{*}}  &  \multicolumn{8}{c}{Trigger type}  \\ \cline{4-11}
   &   method &     & None &Gaussian &Color & Filter &  Patch & Blend & Refool  & UAP\\
\midrule

 \multirow{12}{*}{DPIR} 
& \multirow{3}{*}{SW}  & Normal.   & 0.8381  &0.7915    &  0.7812 &  0.6328  &  0.7320& 0.7259 &0.7531   & 0.7920  \\
&   & Effect.  &  $\backslash$ &  0.9475  & 0.4166  &   0.5466&  0.9864& 0.9618 & 0.6862 & 0.9631\\ 
&   & Total  &  $\backslash$ & 1.7390  & 1.1978   & 1.1794  &  1.7184&   1.6877& 1.4393 &\textbf{1.7551} \\ \cline{2-11}
 & \multirow{3}{*}{DWA} & Normal.   & $\backslash$  & 0.7884  &  0.5442  &   0.6739 &  0.7937 &  0.7641 &0.7517  & 0.7866  \\
&   & Effect.  &  $\backslash$ &   0.9954 & 0.7099  &  0.5146 &  0.6529&  0.8448&  0.8982& 0.9763\\ 
&   & Total  &  $\backslash$ & \textbf{1.7838}   &  1.2541 &  1.1885 &  1.4466& 1.6089 &   1.6499& 1.7629 \\ \cline{2-11}
 &  \multirow{3}{*}{UW} & Normal.  &$\backslash$  & 0.7925  &0.6956   & 0.7066  &0.7694  & 0.7870  &  0.8154  &  0.8308 \\
&   & Effect.  &  $\backslash$ &  0.9953 &0.5773   & 0.6134  & 0.9093 & 0.9669  & 0.9228  &0.9887 \\ 
&   & Total  &  $\backslash$ & 1.7878   &  1.2729  &  1.3200 &  1.6787& 1.7539 & 1.7382 &  \textbf{1.8195}\\ \cline{2-11}
&  \multirow{3}{*}{PCGrad} & Normal.   &$\backslash$  & 0.7966  &  0.7428  &  0.6616 & 0.7818 & 0.7933  & 0.8218  & 0.8201  \\
&   & Effect.  &  $\backslash$ &  0.9938 & 0.5981  & 0.5387  &  0.8456&  0.9734&  0.9143& 0.9862\\ 
&   & Total  &  $\backslash$ & 1.7904   &  1.3409  &  1.2003  &  1.6274 &   1.7667& 1.7361 & \textbf{1.8063}\\  \cline{1-11}

\multirow{12}{*}{SCUNet} 
& \multirow{3}{*}{SW}  & Normal.   & 0.8492  & 0.7912   & 0.8082  &  0.7738  &0.8092   &0.6438   & 0.7876  & 0.8476\\
&   & Effect.  &  $\backslash$ &  0.8678  &   0.6631 &  0.6199 & 0.8243 &0.8059  &0.5305  &0.8615 \\ 
&   & Total  &  $\backslash$ & 1.6590  &1.4713    &  1.3937 & 1.6335 &1.4497  & 1.3181 &  \textbf{1.7091} \\ \cline{2-11}
 & \multirow{3}{*}{DWA} & Normal.   & $\backslash$  & 0.7367 & 0.6284   &0.8262   &0.8227 & 0.7201  &0.7818  &  0.8227\\
&   & Effect.  &  $\backslash$ &  0.9594 &  0.8914 &  0.7317 &0.7255  & 0.8473 &   0.7597&0.8971 \\ 
&   & Total  &  $\backslash$ &  1.6961   & 1.5198  &  1.5579  & 1.5482  & 1.5673 & 1.5415 &  \textbf{1.7198}\\ \cline{2-11}
 &  \multirow{3}{*}{UW} & Normal.   &$\backslash$  & 0.7445   & 0.7239   & 0.6484  &0.7798  & 0.6958  & 0.7604  &  0.8285  \\
&   & Effect.  &  $\backslash$ &0.9717   & 0.9933  &0.8445   & 0.8954 &  0.8480 &0.7703  & 0.9075\\ 
&   & Total  &  $\backslash$ & 1.7162  &  1.7172 & 1.4929   & 1.6752  & 1.5438 & 1.5307  & \textbf{1.7360}\\ \cline{2-11}
&  \multirow{3}{*}{PCGrad} & Normal.   &$\backslash$  & 0.7520 & 0.7202  & 0.6370  & 0.7921  & 0.7341 & 0.6704   &0.8357  \\
&   & Effect.  &  $\backslash$ & 0.9379  & 0.9930  &  0.8918 &0.8272  &0.8892  &0.8982  &0.8750 \\ 
&   & Total  &  $\backslash$ &  1.6899   &  \textbf{1.7132}  &  1.5288  &1.6193  &  1.6233 &  1.5686& 1.7107\\  \cline{1-11}

\multirow{12}{*}{MPRNet} 
& \multirow{3}{*}{SW}  &Normal.   & 0.8737  &  0.8536   & 0.7683  & 0.7467   & 0.7401  &0.7030  & 0.7890   & 0.8732  \\
&   & Effect.  &  $\backslash$ & 0.9855  &   0.3369 & 0.2216  & 0.4000 &0.5619  & 0.4922  & 0.9736 \\ 
&   & Total  &  $\backslash$ &1.8391   & 1.1052   & 0.9683  & 1.1401 &1.2649   & 1.2812 & \textbf{1.8468}\\ \cline{2-11}
 & \multirow{3}{*}{DWA} & Normal.   & $\backslash$  & 0.8019 & 0.8167  &0.6647    &  0.8696 &0.7529 &0.8431  &0.8729  \\
&   & Effect.  &  $\backslash$ & 0.9638  & 0.2939  &  0.4193 & 0.5048 &  0.7806&  0.7711 &  0.9832\\ 
&   & Total  &  $\backslash$ &   1.7657  &  1.1106 &  1.0840 &  1.3744 &  1.5335   & 1.6142  &\textbf{1.8561}  \\ \cline{2-11}
 &  \multirow{3}{*}{UW} &Normal.  &$\backslash$  &  0.8085   & 0.8494 &0.7961    & 0.8631 &0.7705  & 0.8212   &0.8745 \\
&   & Effect.  &  $\backslash$ & 0.9900  &0.3532   &  0.2410  & 0.5041 & 0.6751 & 0.7973 & 0.9910\\ 
&   & Total  &  $\backslash$ & 1.7985  & 1.2026    &  1.0371 & 1.3672   &  1.4456 &1.6185  & \textbf{1.8655}   \\ \cline{2-11}
&  \multirow{3}{*}{PCGrad} & Normal.   &$\backslash$  &   0.8085  & 0.8341   & 0.7659   & 0.8715  & 0.6517 & 0.8418 &0.8719   \\
&   & Effect.  &  $\backslash$ &  0.9779 & 0.3902  & 0.2305  &   0.4988 &0.6036  &0.7459  & 0.9879\\ 
&   & Total  &  $\backslash$ & 1.7864  & 1.2243  & 0.9964  & 1.3703 &  1.2553 &  1.5877  &  \textbf{1.8598}\\  \cline{1-11}

\multirow{12}{*}{MIRNet} 
& \multirow{3}{*}{SW}  &Normal.  &0.8673   &  0.7423  & 0.6467 &  0.7464  &  0.8700  & 0.7459  &0.6828  &0.8664  \\
&   & Effect.  &  $\backslash$ & 0.9497  &  0.8711 &   0.5154 &  0.9951 &0.2901  &  0.5696& 0.9844\\ 
&   & Total  &  $\backslash$ &1.6920    &  1.5178   &1.2618   & 1.8651 & 1.0360 &  1.2524&\textbf{1.8508} \\ \cline{2-11}
 & \multirow{3}{*}{DWA} & Normal.  & $\backslash$  &  0.8688  & 0.5337 &  0.6440 &0.8668 &   0.7802 &0.6921 & 0.8646 \\
&   & Effect.  &  $\backslash$ & 0.9779  & 0.7573  & 0.9011  &0.9961  & 0.7827 &  0.4265 & 0.9968\\ 
&   & Total  &  $\backslash$ &  \textbf{1.8667}  &  1.2910  &  1.5451   &  1.8629  & 1.5629 &  1.1186  &1.8614  \\ \cline{2-11}
 &  \multirow{3}{*}{UW} &Normal.   &$\backslash$  & 0.7312   & 0.8404   &   0.7889 &  0.8692  & 0.6879  &0.7393  &  0.8705\\
&   & Effect.  &  $\backslash$ &  0.9793  &   0.8187& 0.7569  &0.9946  & 0.5735 & 0.9041 & 0.9990 \\ 
&   & Total  &  $\backslash$ &  1.7105 & 1.6591  &  1.5458 & 1.8638 &1.2614  & 1.6434  &  \textbf{1.8695} \\ \cline{2-11}
&  \multirow{3}{*}{PCGrad} & Normal.   &$\backslash$  & 0.8047  & 0.8320  &0.6444  & 0.8648 & 0.6860  & 0.6751 & 0.8692 \\
&   & Effect.  &  $\backslash$ & 0.9706  &   0.8689 & 0.9236  & 0.9956 &0.3490  &   0.8780& 0.9949\\ 
&   & Total  &  $\backslash$ & 1.7753   &1.7009    &  1.5680  &1.8604  &  1.0350& 1.5531  & \textbf{1.8641}\\  \cline{1-11}

\multirow{12}{*}{ESRGAN} 
& \multirow{3}{*}{SW}  & Normal.   &  0.8650 & 0.8735  & 0.8186  & 0.8005  &  0.8718 & 0.8535  &0.8228  &0.8713  \\
&   & Effect.  &  $\backslash$ & 0.9818  &  0.3929  & 0.6037  &0.2934  & 0.4067 & 0.3409 & 0.9950 \\ 
&   & Total  &  $\backslash$ &1.8553    &  1.2115  & 1.4042   &  1.1652 &1.2602   &1.1637   & \textbf{1.8663}\\ \cline{2-11}
 & \multirow{3}{*}{DWA} & Normal.   & $\backslash$  & 0.8719  &  0.7592  &   0.6913 &0.8182   &0.8277 &  0.8032  & 0.8690 \\
&   & Effect.  &  $\backslash$ &  0.9941 & 0.6040  & 0.5489  &0.3594  &0.4416  &0.3585  & 0.9971 \\ 
&   & Total  &  $\backslash$ & 1.8660   &   1.3632&  1.2402  & 1.1776 &  1.2693  &1.1617  & \textbf{1.8661} \\ \cline{2-11}
 &  \multirow{3}{*}{UW} &Normal.   &$\backslash$  & 0.8656   & 0.5908  & 0.8433   & 0.8640 & 0.8349 &  0.8314  &  0.8686   \\
&   & Effect.  &  $\backslash$ &  0.9946  &  0.8538 &0.6614   &0.1596  &  0.7015 &0.4340  & 0.9936\\ 
&   & Total  &  $\backslash$ & 1.8602  &1.4446    & 1.5047  &1.0236   & 1.5364  &  1.2654 & \textbf{1.8622}\\ \cline{2-11}
&  \multirow{3}{*}{PCGrad} & Normal.  &$\backslash$  &  0.8752  & 0.7274   & 0.8112   & 0.8546   &0.8460 &  0.8113   & 0.8728   \\
&   & Effect.  &  $\backslash$ &0.9907   & 0.7271  &  0.7494 &0.2688  &0.5862  &  0.4314& 0.9969 \\ 
&   & Total  &  $\backslash$ & 1.8659  &  1.4545 &  1.5606 & 1.1234 &  1.4322  & 1.2427 &\textbf{1.8697} \\ 

\bottomrule
\end{tabular}
 \begin{tablenotes}
        \footnotesize
        \item[*] Normal. denotes the normal-functionality; Effect. denotes the effectiveness; Total represents the sum of them. The bolded results represent the maximum total score.
      \end{tablenotes}
    \end{threeparttable}}
\vspace{-10pt}
\end{table*}

We have conducted extensive experiments of I2I backdoor attacks with different backdoor triggers and MTL methods on various I2I network architectures. As presented in Table \ref{Table: Performance of I2I backdoor with different backdoor triggers and MTL methods on image denoising task.}, most triggers achieve high attack effectiveness in attacking image denoising task. However, they fail to preserve the similar normal-functionality with that of the clean model. In comparison, the UAP trigger is superior to other triggers in maintaining normal-functionality. As provided in Table \ref{Table: Performance of I2I backdoor with different backdoor triggers and MTL methods on image super-resolution task.}, only the UAP trigger achieves good attack performance on all these I2I models. In addition, we also calculate the sum of the normal-functionality and the attack effectiveness to see which trigger achieves a better balance of preserving normal-functionality and enhancing attack effectiveness. The results show that the UAP trigger always achieves the highest total score on all considered I2I network architectures.


\subsubsection{Computational Overhead}

We have assessed the computational overhead of generating the UAP trigger. As provided in Table \ref{Table: Computational overhead for trigger generation}, the computational overhead of the UAP trigger generation algorithm is relatively small and falls within acceptable bounds for potential backdoor attackers. Hence, in the subsequent experiments, we use the UAP trigger to perform our I2I backdoor attack.

\begin{table}[H]\small
\begin{center}
\caption{Computational overhead (s) for the UAP trigger generation.}
\label{Table: Computational overhead for trigger generation}
\begin{tabular}{ccccc}
\toprule
DPIR & SCUNet & MPRNet  & MIRNet &ESRGAN\\
\hline
18.08  & 49.54 & 52.22&128.13& 57.64 \\
\bottomrule
\end{tabular}
\end{center}
\end{table}


Furthermore, we have also evaluated the computational overhead of different MTL methods. As outlined in Table \ref{Table: Computational overhead for different MTL methods}, the difference between the computational overhead of these MTL methods is relatively negligible. Therefore, without loss of generality, we have opted to employ the PCGrad method for MTL in the subsequent experiments.

\begin{table}[H]\small
\begin{center}
\caption{Computational overhead (s) for different MTL methods (1 epoch).}
\label{Table: Computational overhead for different MTL methods}
\resizebox{0.98\linewidth}{!}{
\begin{tabular}{c|cccc}
\toprule
\diagbox[width=10em,trim=l]{Architecture}{MTL method} & SW & UW & DWA& PCGrad  \\
\hline
DPIR & 13.64 &  8.93  &  14.15 & 16.58   \\
SCUNet  & 31.84 & 25.95 & 30.15 & 39.01\\
MPRNet  &  35.39&36.91  &33.66  &43.52 \\
MIRNet  &  93.04  &  85.09  & 77.40 &   85.43 \\
ESRGAN  & 54.72 & 56.11 & 60.47 &71.86 \\

\bottomrule
\end{tabular}}
\end{center}
\end{table}

\subsection{Ablation Study}
\subsubsection{Ablation Study of the Loss Function}
In this section, instead of using SSIM metric and MSE loss, we employ the perceptual loss function and use LPIPS \cite{zhang2018unreasonable} as the metric. LPIPS uses pre-trained deep networks to extract image features and then calculates the distance between these features to evaluate the perceptual similarity between images. Specifically, we replace the main task loss and backdoor task loss with the perceptual loss functions based on LPIPS metric: $\mathcal{L}_{m} = lpips(F(X_n),Y_n)$, $\mathcal{L}_{b} = lpips(F(X_b),Y_b)$. The $lpips(x,\hat{x})$.

After that, the UAP trigger and PCGrad MTL method are selected as an example\footnote{Other cases yield the same experimental results.} to evaluate the attack performance across different architectures. As presented in Table \ref{Table: attack performance for LPIPS metric}, the ``Normal.'' of clean/backdoor model denotes the LPIPS distance of the denoised (or super-resolution) image and the original image, representing the normal-functionality of the clean/backdoor model; the ``Effect.'' of backdoor model denotes the LPIPS distance of the denoised (or super-resolution) version of the triggered image and the backdoor target image, representing the attack effectiveness of the backdoor model. It can be observed that I2I backdoor does not cause significant degradation in the normal-functionality of the backdoor model. Besides, it also achieves high attack effectiveness, i.e., the denoised (or super-resolution) version of the triggered image is extremely similar to the backdoor target image. Therefore, it can be concluded that I2I backdoor is highly generalisable and can achieve great attack performance for different loss functions and similarity metrics.

\begin{table}[H]\small
\begin{center}
\caption{The attack performance of I2I backdoor with the perceptual loss functions based on LPIPS metric.}
\label{Table: attack performance for LPIPS metric}
\resizebox{0.98\linewidth}{!}{
\begin{tabular}{cc|cc|cc}
\toprule
\multirow{2}{*}{Task} & \multirow{2}{*}{Architecture} & \multicolumn{2}{c|}{Clean model} &  \multicolumn{2}{c}{Backdoor model}  \\ \cline{3-6}
 &  & Normal. & Effect. & Normal. & Effect.  \\
\hline
\multirow{5}{*}{Denoise}&DPIR & 0.0217 & $\backslash$  &  0.0228 & 0.0029   \\
&SCUNet  & 0.0261 & $\backslash$ &0.0270 & 0.0070\\
&MPRNet  & 0.0289 &$\backslash$  &0.0295  &0.0066 \\
&MIRNet  & 0.0252  &  $\backslash$  & 0.0258 &   0.0019 \\
&ESRGAN  &0.0318 & $\backslash$ & 0.0335 &0.0025 \\ \cline{1-6}
\multirow{5}{*}{SR}&DPIR & 0.0752 & $\backslash$  &  0.0770 & 0.0051   \\
&SCUNet  & 0.0912 & $\backslash$ & 0.0933 & 0.0092\\
&MPRNet  &  0.0809&$\backslash$  &0.0838  &0.0083 \\
&MIRNet  &  0.0834  &  $\backslash$  & 0.0882 &   0.0045 \\
&ESRGAN  & 0.0887 & $\backslash$ & 0.0901 &0.0058 \\
\bottomrule
\end{tabular}}
\end{center}
\end{table}

\subsubsection{Impact of Different Weights on Attack Performance}
\begin{figure}
  \centering
  \subfigure[Denoise task]{\includegraphics[width=0.45\linewidth]{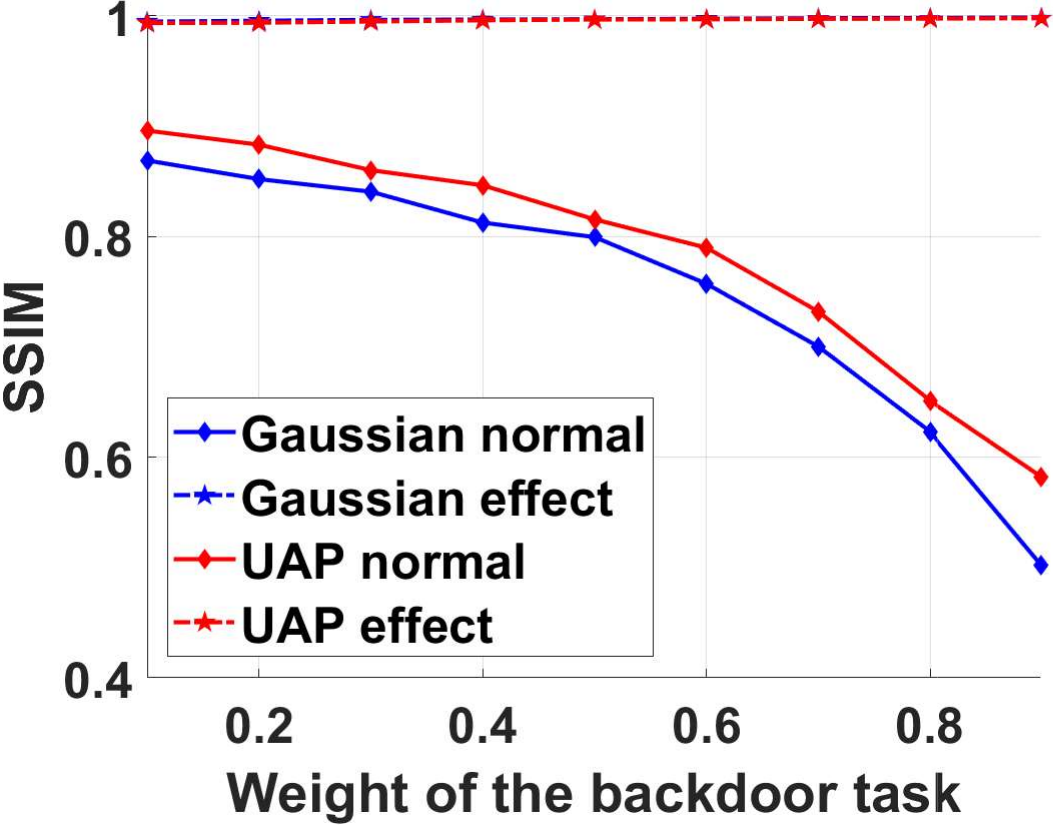}}
  \subfigure[Super-resolution task]{\includegraphics[width=0.45\linewidth]{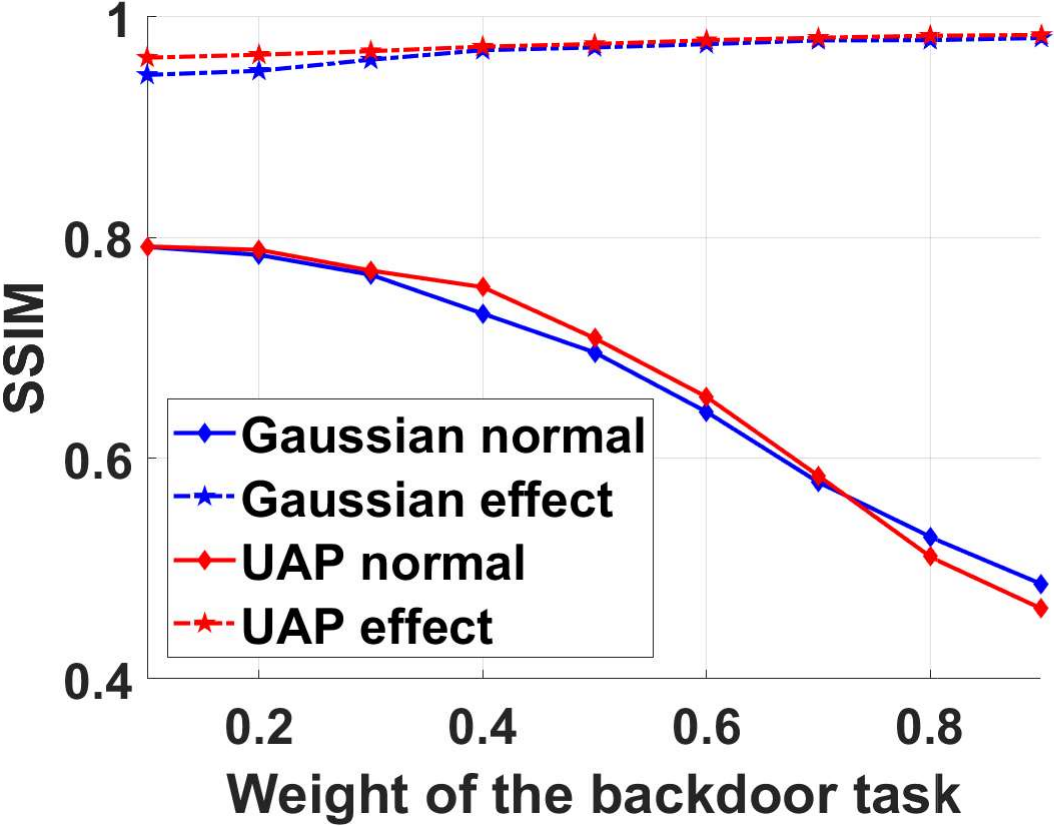}}
\caption{The attack performance under different weights.}
\label{Fig:The attack performance under different weights}
\end{figure}

In this section, we conduct experiments to evaluate the attack performance under different weights for main task and backdoor task. Since the weights assigned by dynamic weighting methods (UW, DWA, and PCGrad) are constantly changing, we mainly consider the static weighting method. Specifically, we select the Gaussian and UAP trigger on DPIR model as an example, set the weight of the backdoor task from 0.1 to 0.9 and present the experimental results in Figure \ref{Fig:The attack performance under different weights}. It can be seen that as the weight of the backdoor task increases, the model normal-functionality decreases dramatically; and the increase in the weight of the backdoor task has no significant improvement in attack effectiveness. Therefore, in our default settings, we set the weight of the main task to 0.9 and the weight of the backdoor task to 0.1 in the static weighting method.

\subsubsection{Impact of Different MTL Weighting Methods on Convergence Rates}

\begin{figure*}
  \centering
  \subfigure[DPIR]{\includegraphics[width=0.18\linewidth]{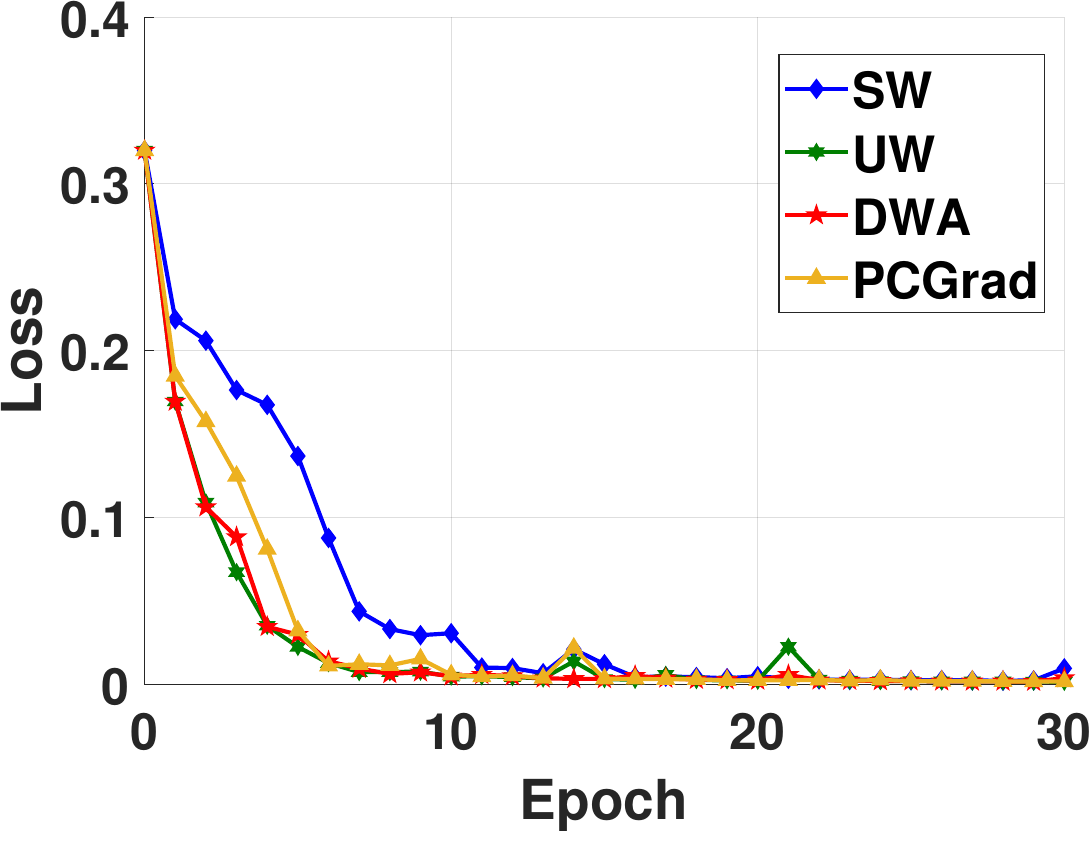}}
  \subfigure[SCUNet]{\includegraphics[width=0.18\linewidth]{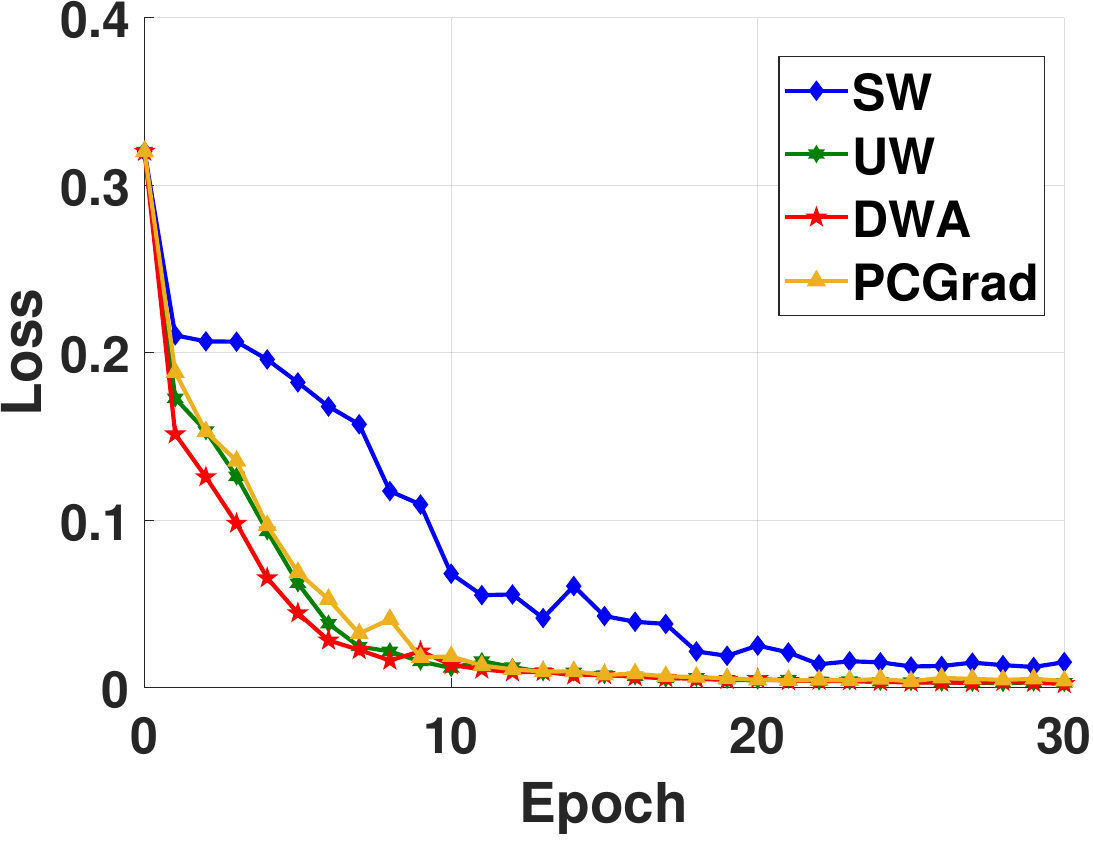}}
  \subfigure[MPRNet]{\includegraphics[width=0.18\linewidth]{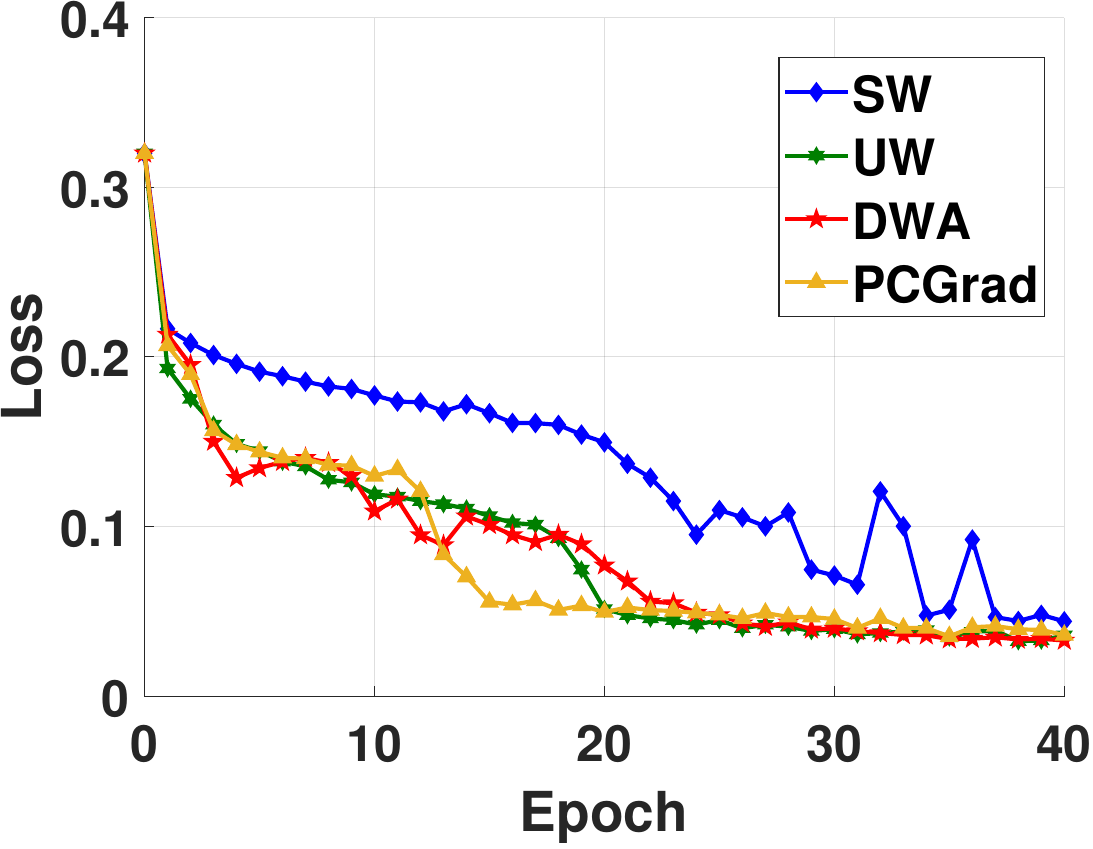}}
  \subfigure[MIRNet]{\includegraphics[width=0.18\linewidth]{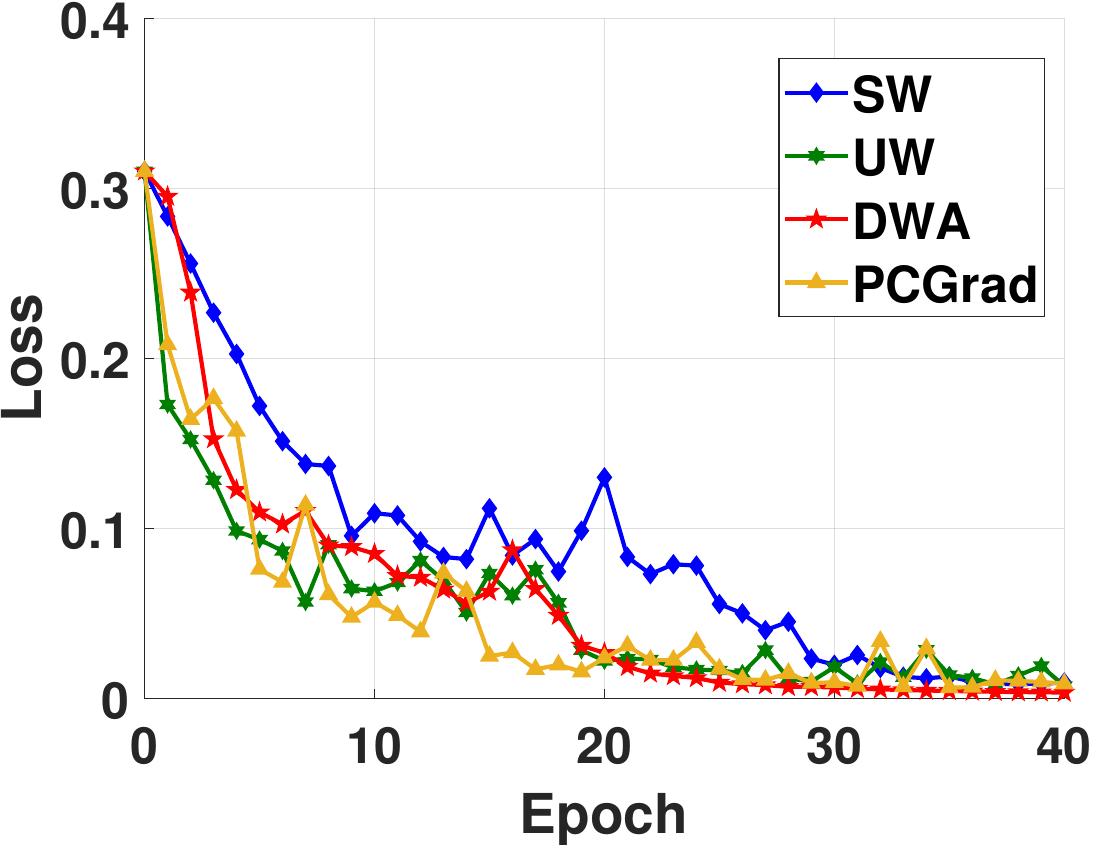}}
  \subfigure[ESRGAN]{\includegraphics[width=0.18\linewidth]{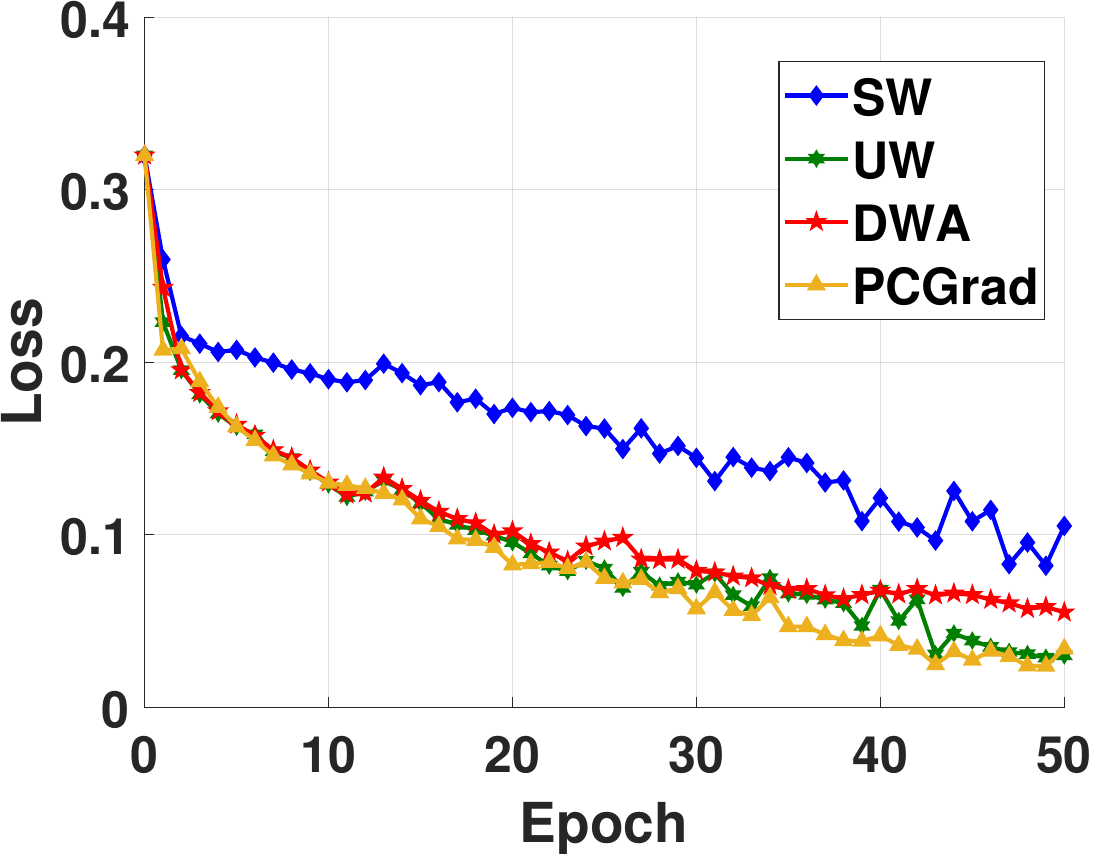}}
\caption{The convergence rates of the training loss with different MTL methods.}
\label{Fig:The convergence rates of the training loss with different MTL methods}
\end{figure*}

We further carry out thorough ablation studies focused on the impact of different MTL weighting methods on the convergence rate. As illustrated in Figure \ref{Fig:The convergence rates of the training loss with different MTL methods}, it can be observed that the dynamic weighting methods, including UW, DWA, and PCGrad, always outperform the static weighting method in terms of convergence rates. This phenomenon can be attributed to the inherent complexity of the I2I backdoor task, which involves mapping a triggered image to an entirely unrelated backdoor target image. Such complexity invariably leads to conflicts with the main task. The static weighting method tends to struggle to achieve an optimal balance between these competing tasks, resulting in reduced training efficiency. Hence, the dynamic weight methods emerge as the more sensible choice for facilitating the I2I backdoor training process.

\subsection{Stealthiness Evaluation}
\label{sec:Stealthiness Evaluation}

In addition to evaluating the effect of backdoor injection on the decrease of model normal-functionality, we further calculated the LPIPS distance of the backdoor-triggered images from the original images to assess the stealthiness of the attack. It can be seen from Table \ref{Table: Stealthiness evaluation of different trigger types and datasets} that the backdoor-triggered images maintain high stealthiness, the LPIPS distances are all below 0.08. The patch trigger is the most stealthy trigger type with the metric of LPIPS. This may be because LPIPS is less sensitive to local, meaningless perturbations.

\begin{table}[H]\small
\begin{center}
\caption{Stealthiness evaluation of different trigger types and testing datasets.}
\label{Table: Stealthiness evaluation of different trigger types and datasets}
\resizebox{0.98\linewidth}{!}{
\begin{tabular}{c|cccc}
\toprule
\diagbox[width=8em,trim=l]{Trigger}{Dataset} & CSet8 & Set14 & ImageNet& Pascal VOC  \\
\hline
Gaussian & 0.0568 & 0.0506   &  0.0714 & 0.0785   \\
Color  & 0.0087  & 0.0257 & 0.0229 & 0.0198\\
Filter  &  0.0195 &0.0381  &0.0458  &0.0316 \\
Patch  &  0.0071 & 0.0041 & 0.0040 &0.0051 \\
Blend  &  0.0328  &  0.0107  & 0.0181 &   0.0202 \\
Refool  & 0.0134 & 0.0327 & 0.0427 &0.0294 \\
UAP  &  0.0317  &  0.0342  &  0.0413  &  0.0280   \\
\bottomrule
\end{tabular}}
\end{center}
\end{table}

\subsection{Robustness Evaluation}
\label{sec:Robustness Evaluation}
In this section, we turn our attention to the robustness evaluation of the I2I backdoor attack against various defense methods. It should be pointed out that many backdoor defense techniques are designed for neural network classifiers, such as Neural Cleanse \cite{wang2019neural}, STRIP \cite{gao2019strip}, and Spectral Signature \cite{NEURIPS2018_280cf18b}, they are not directly applicable to our I2I backdoor attacks. Thus, we have selected three defense methods, including bit depth reduction \cite{xufeature}, image compression \cite{xue2022compression} and model fine-tuning to evaluate the robustness of the I2I backdoor attacks.

\begin{figure*}
  \centering
  \subfigure[DPIR]{\includegraphics[width=0.18\linewidth]{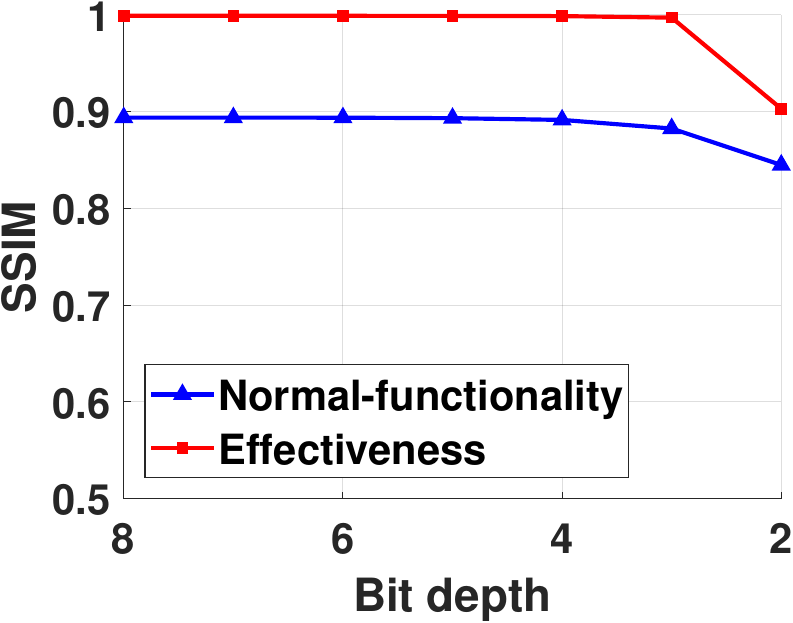}}
  \subfigure[SCUNet]{\includegraphics[width=0.18\linewidth]{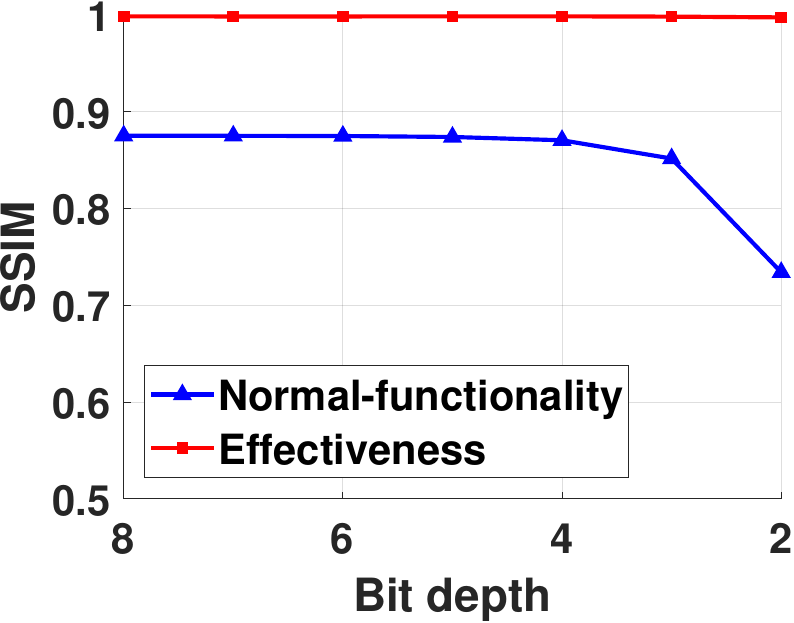}}
  \subfigure[MPRNet]{\includegraphics[width=0.18\linewidth]{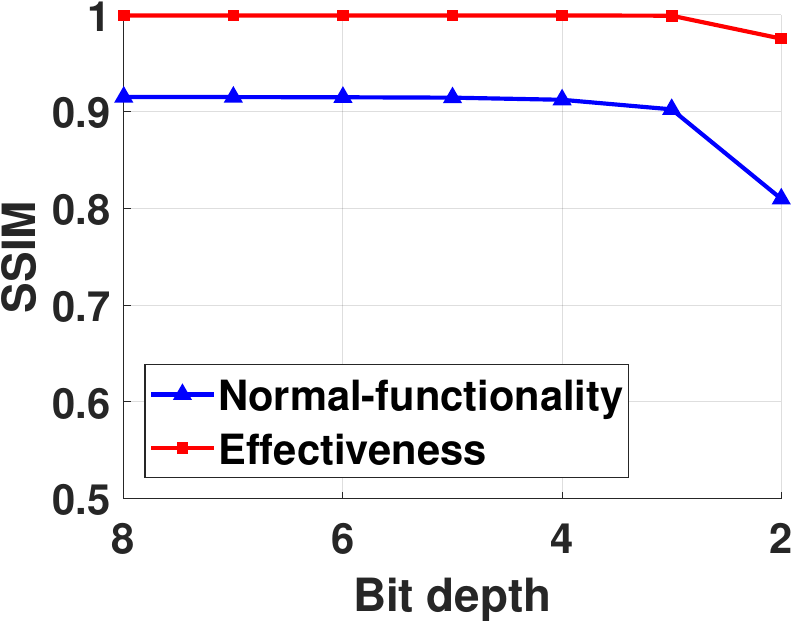}}
  \subfigure[MIRNet]{\includegraphics[width=0.18\linewidth]{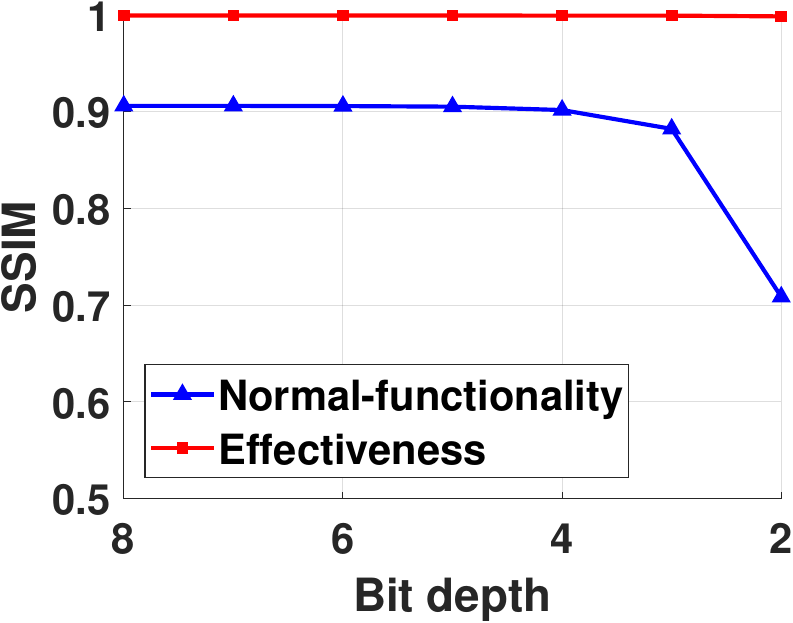}}
  \subfigure[ESRGAN]{\includegraphics[width=0.18\linewidth]{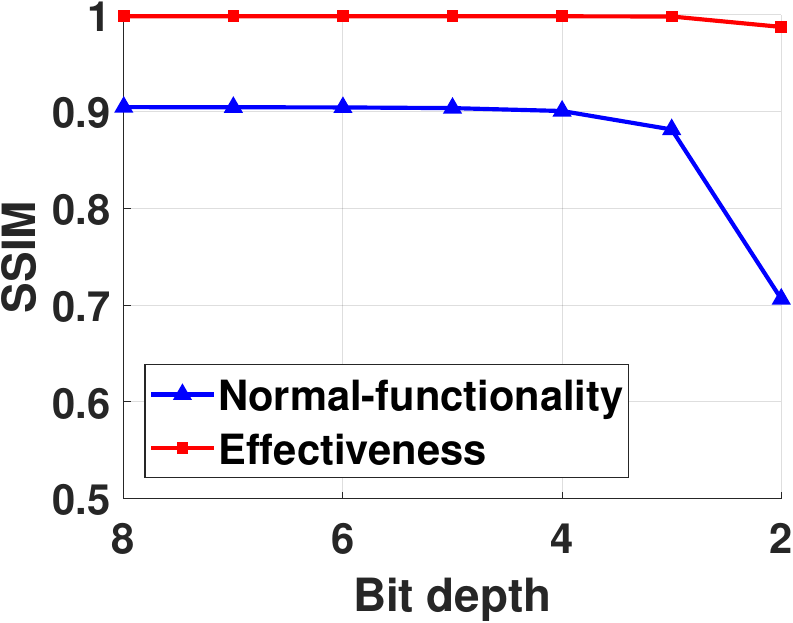}}
\caption{The performance of I2I backdoor attack under bit depth reduction.}
\label{Fig:The performance of I2I backdoor attack under bit depth reduction.}
\end{figure*}

\textbf{Bit depth reduction.} We reduce the bit depth of input images before sending them to I2I models. As illustrated in Figure \ref{Fig:The performance of I2I backdoor attack under bit depth reduction.}, the effectiveness of the attack consistently maintains a high level as the bit depth decreases. It demonstrates that the preprocessing of bit depth reduction is ineffective in mitigating our I2I backdoor attack.

\begin{figure*}
  \centering
  \subfigure[DPIR]{\includegraphics[width=0.18\linewidth]{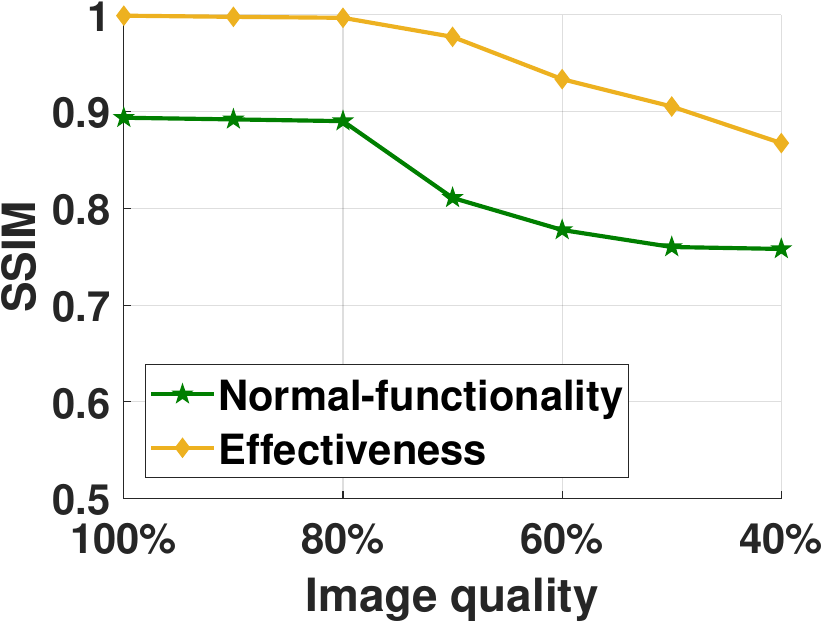}}
  \subfigure[SCUNet]{\includegraphics[width=0.18\linewidth]{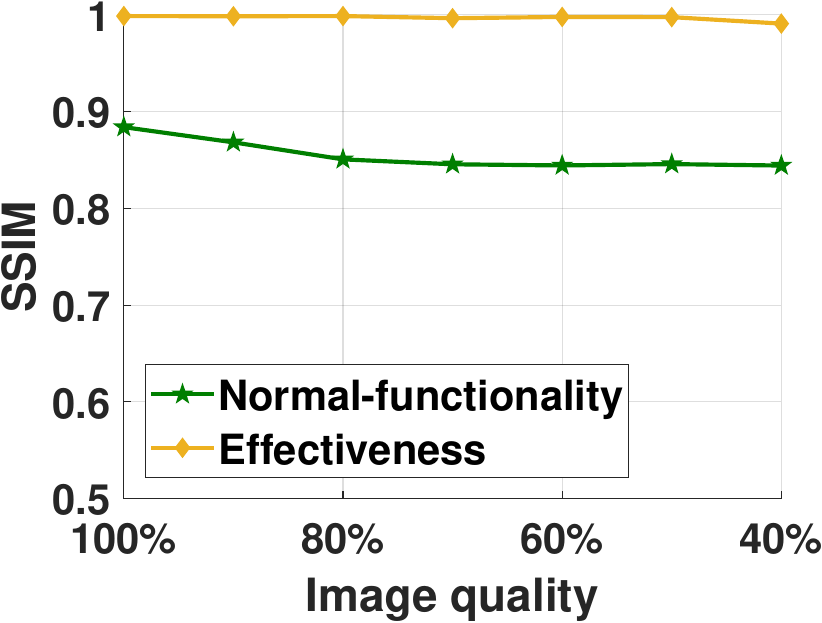}}
  \subfigure[MPRNet]{\includegraphics[width=0.18\linewidth]{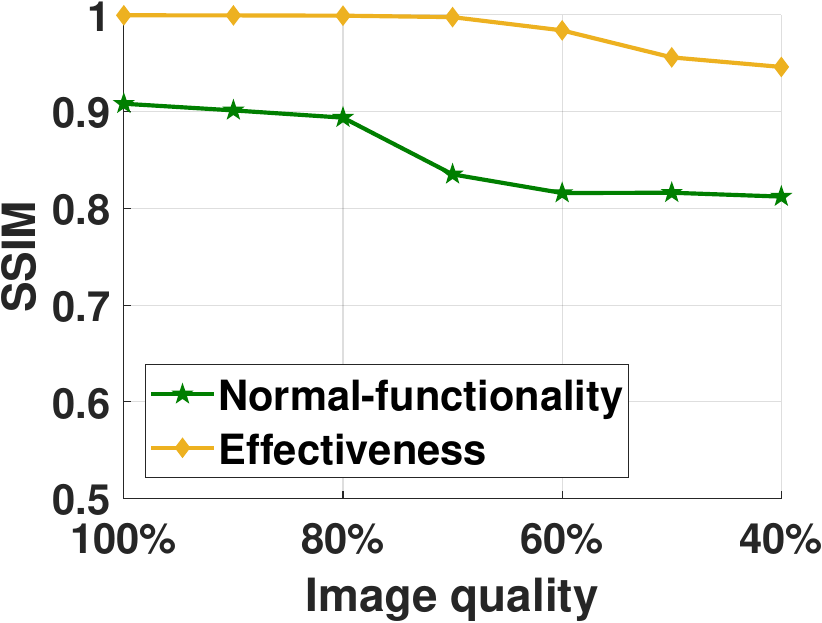}}
  \subfigure[MIRNet]{\includegraphics[width=0.18\linewidth]{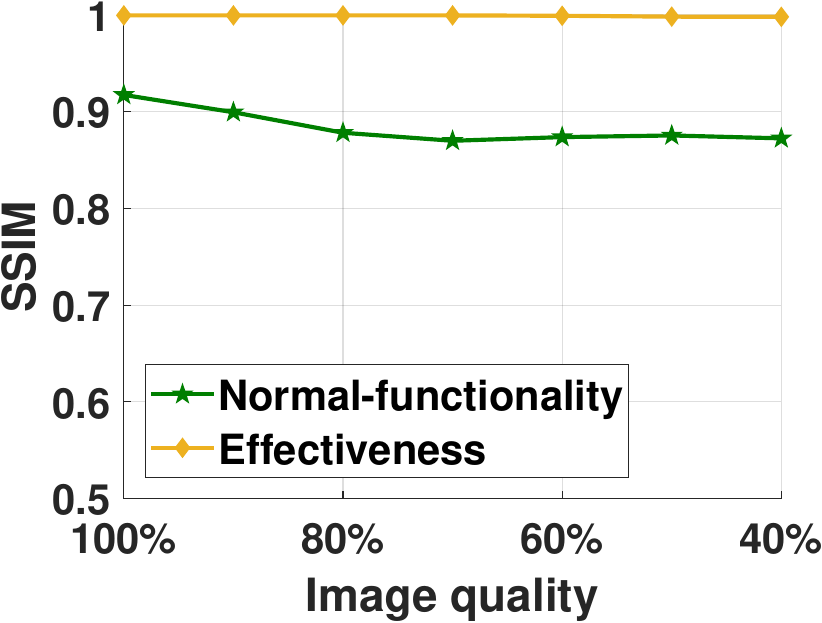}}
  \subfigure[ESRGAN]{\includegraphics[width=0.18\linewidth]{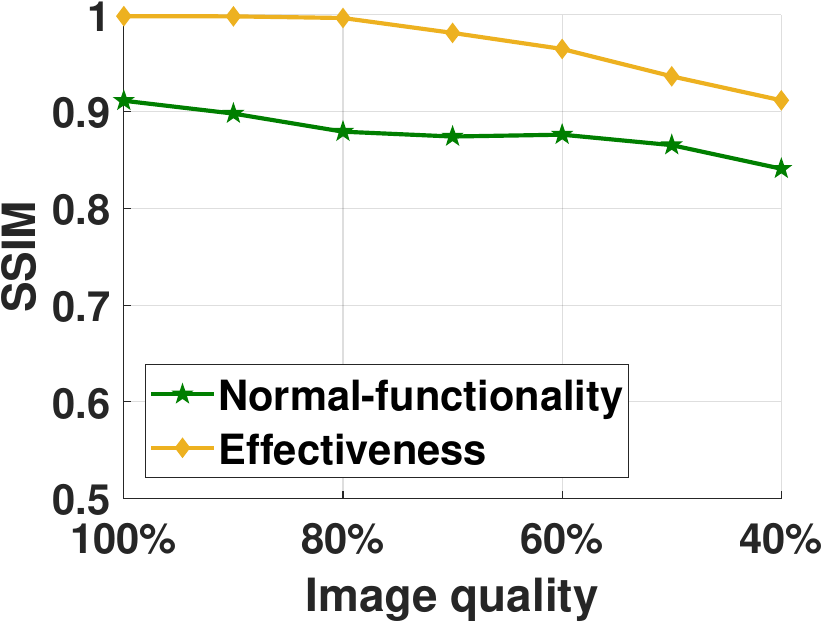}}
\caption{The performance of I2I backdoor attack under image compression.}
\label{Fig:The performance of I2I backdoor attack under image compression.}
\end{figure*}

\textbf{Image compression.}  We compress input images before sending them to I2I models. As depicted in Figure \ref{Fig:The performance of I2I backdoor attack under image compression.}, the degradation in normal-functionality consistently outweighs the degradation in attack effectiveness as input images undergo image compression. Thus, the preprocessing of image compression is also far from an effective defense method against the proposed I2I backdoor attack.

\textbf{Model fine-tuning.} We assume that the defender has a small amount\footnote{In our experiments, this amount is assumed to be 10\% of the original training dataset.} of clean images and uses these images to fine-tune the backdoored I2I model. As presented in Table \ref{Table: complete SONG and SOTB for I2I backdoor attack under model fine-tuning}, the I2I backdoor remains effective after fine-tuning with clean images.

\begin{table}\small
\begin{center}
\tabcolsep=0.1cm
\caption{Performance of I2I backdoor attack under model fine-tuning.}
\label{Table: complete SONG and SOTB for I2I backdoor attack under model fine-tuning}
\resizebox{0.98\linewidth}{!}{
\begin{tabular}{cc|ccccc}
\toprule
Epoch &SSIM &DPIR & SCUNet & MPRNet & MIRNet &ESRGAN\\
\hline 
\multirow{2}{*}{10} & Normal. &  0.8939 &  0.8855  &  0.9145   &0.9049 &0.9038  \\
  & Effect. &  0.9990 &  0.9989  &  0.9995 &0.9994 & 0.9978 \\  \hline
\multirow{2}{*}{20} & Normal. &  0.8935 &  0.8850  &  0.9156 & 0.9056& 0.9039\\
& Effect. &  0.9991 &  0.9981  &  0.9994  &0.9994 & 0.9987\\ \hline
\multirow{2}{*}{30} & Normal. &  0.8940 &  0.8857  &  0.9162 & 0.9052&0.9036 \\
& Effect.  &  0.9987 & 0.9986  &  0.9992  &0.9983 & 0.9983\\ \hline
\multirow{2}{*}{40} & Normal. &  0.8944 &  0.8853  &  0.9160  &0.9056 &0.9034 \\
& Effect.&  0.9989 &  0.9992  &  0.9988  &0.9989 &0.9987 \\ \hline
\multirow{2}{*}{50}& Normal.  & 0.8948 &  0.8851  &  0.9161 &0.9058 &0.9037 \\
& Effect.& 0.9985 &  0.9980  &  0.9990  &0.9994 & 0.9985 \\ 
\bottomrule
\end{tabular}}
\end{center}
\end{table}

\begin{figure}
\centering
\includegraphics[width=0.35\textwidth]{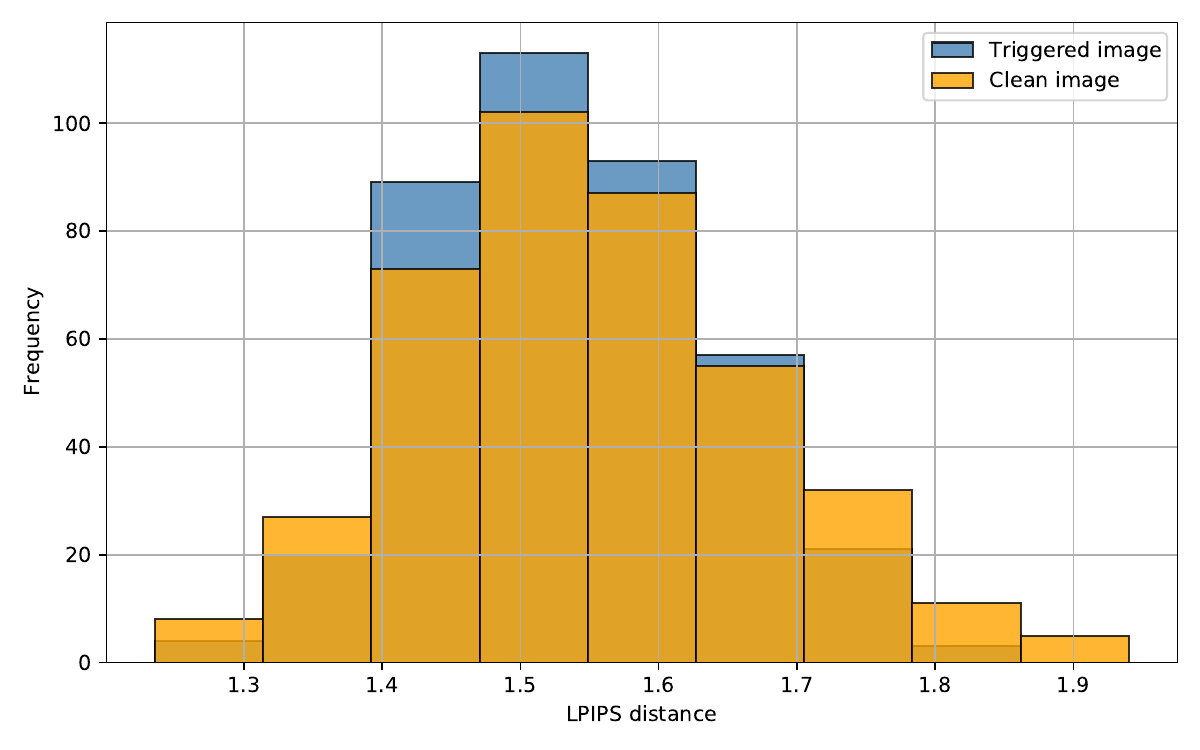}
\caption{Robustness of I2I backdoor under STRIP.}
\label{Fig:Robustness of I2I backdoor under STRIP}
\end{figure}

\textbf{STRIP.} We extend the representative backdoor detection method STRIP \cite{gao2019strip} to I2I tasks and evaluate the robustness of the I2I backdoor under this defense. STRIP is a testing-time backdoor detection method that detects whether a testing image contains a trigger. The idea of STRIP is based on the assumption that the backdoor trigger is robust and can remain effective when a triggered image is superimposed by a clean image. In our work, we design the STRIP method for the I2I task following the framework of the STRIP method in the image classification task. Specifically, we sample 400 clean images and superimpose the target image with these clean images separately, and send them to I2I models to obtain output images. After that, we calculate the LPIPS distances between the output image corresponding to each superimposed image and the output image corresponding to the original target image. If the target image is backdoor-triggered, the output images will show high similarity. Conversely, if the target image is clean, the output images will show low similarity. Hence, we select the UAP trigger on DPIR denoise model as an example, and calculate and compare the LPIPS distance distributions of the clean and triggered image. As illustrated in Figure \ref{Fig:Robustness of I2I backdoor under STRIP}, we observe that the clean and triggered target image have very similar LPIPS distance distributions so that STRIP is not able to distinguish whether a testing image contains the backdoor trigger or not. This is mainly because the UAP trigger is destroyed when the triggered image is superimposed with a clean image, the backdoor trigger becomes ineffective in the superimposed image.

\subsection{Evaluation on I2I Backdoor Attack against Downstream Classification Task}
\label{sec:Evaluation on I2I Backdoor Attack against Downstream Classification Task}

To perform the I2I backdoor attack against the downstream classification task, we employ the Algorithm \ref{alg:The Generation Algorithm of UAP against classification models} to generate the UAP against the pre-trained ResNet152 classifier. After that, we employ this UAP to embed the I2I backdoor attack into the upstream image denoising model. Finally, we evaluate the attack performance on other clean pre-trained classifiers (including ResNet50, VGG19 and MobileNetV2) to measure the attack transferability. 

In the case of the I2I backdoor attack against the downstream object detection task, we employ the UAP generation algorithm for object detection \cite{chow2020adversarial} to construct the UAP against the pre-trained MobileNetv1-YOLOv3 detector. After that, we employ this UAP to embed the I2I backdoor attack into the upstream image denoising model. Finally, we evaluate the attack performance on other clean pre-trained object detectors (including MobileNetv2-YOLOv3, Darknet53-YOLOv3 and EfficientNet-YOLOv3) to measure the attack transferability.

As presented in Table \ref{Table: The complete performance of I2I backdoor attack against downstream classification task} and \ref{Table: The complete performance of I2I backdoor attack against downstream detection task}, for clean input images, the downstream denoised accuracy/mAP of the backdoor denoising model and the normal denoising model exhibit minimal disparity. This confirms that the I2I backdoor does not affect the normal-functionality of the downstream classification/detection task. In the case of backdoor-triggered input images and the backdoor upstream denoising model, the denoised versions of these images can fool the downstream clean pre-trained classifiers/detectors with high success rates. This proves the attack effectiveness of the I2I backdoor attack against the downstream classification/detection task.

\begin{table}\small
\begin{center}
\tabcolsep=0.1cm
\caption{The performance of I2I backdoor attack against downstream classification task (with the UAP against ResNet152 classifier).}
\label{Table: The complete performance of I2I backdoor attack against downstream classification task}
\resizebox{0.98\linewidth}{!}{
\begin{threeparttable}
\begin{tabular}{cc|cc|c}
\toprule
\multirow{3}{*}{\makecell{Upstream \\ denoising \\ model $D$}} &  \multirow{3}{*}{\makecell{Downstream \\ classification \\ model}}  &   \multicolumn{2}{c|}{Denoised accuracy (\%)}  & ASR (\%)  \\\cline{3-5}
       &              &    Clean $D$   &Backdoor $D$    &Backdoor $D$   \\
       &              &     Clean img.  &  Clean img. &   Backdoor img. \\
\midrule
 \multirow{3}{*}{DPIR} 
& ResNet50    & 72.08 & 71.48&   72.48 \\ \cline{2-5}
 & VGG19   & 65.32 &65.42 &  85.90  \\ \cline{2-5}
&  MobileNetV2   &  64.40   & 64.68  &  74.90  \\ \cline{1-5} 

\multirow{3}{*}{SCUNet} 
& ResNet50    & 71.72 & 71.56&   72.64\\ \cline{2-5}
 & VGG19   & 65.06 & 64.26& 80.96   \\ \cline{2-5}
&  MobileNetV2   &65.66   & 65.20  &  74.74   \\ \cline{1-5} 

\multirow{3}{*}{MPRNet} 
& ResNet50    & 71.34& 71.22&   72.82 \\ \cline{2-5}
 & VGG19   & 64.62& 64.54& 81.14  \\ \cline{2-5}
&  MobileNetV2   & 64.32   & 64.66  & 74.92  \\  \cline{1-5}
 
 \multirow{3}{*}{MIRNet} 
& ResNet50    &   71.64  &  71.40  &  72.74  \\ \cline{2-5}
 & VGG19   &   65.30  &  63.88  & 80.72  \\ \cline{2-5}
&  MobileNetV2   &   65.04  &  64.34  &  75.12  \\  \cline{1-5}

 \multirow{3}{*}{ESRGAN} 
& ResNet50    &  71.16   &   69.80 & 72.78\\ \cline{2-5}
 & VGG19   &    64.42 &  63.36  & 81.48   \\ \cline{2-5}
&  MobileNetV2   &  64.22   & 62.92   & 75.56  \\ 

\bottomrule
\end{tabular}
    \end{threeparttable}}
\end{center}
\end{table}

\begin{table}\small
\begin{center}
\tabcolsep=0.03cm
\caption{The performance of I2I backdoor attack against downstream detection task (with the UAP against MobileNetv1-YOLOv3).}
\label{Table: The complete performance of I2I backdoor attack against downstream detection task}
\resizebox{0.98\linewidth}{!}{
\begin{threeparttable}
\begin{tabular}{cc|cc|c}
\toprule
\multirow{3}{*}{\makecell{Upstream \\ denoising \\ model $D$}} &  \multirow{3}{*}{\makecell{Downstream \\ detection \\ model}}  &   \multicolumn{2}{c|}{mAP (\%)}  & ASR (\%)  \\\cline{3-5}
   &              &    Clean $D$   & Backdoor $D$    &Backdoor $D$   \\
      &              &     Clean img.  & Clean img. &   Backdoor img. \\
\midrule
 \multirow{3}{*}{DPIR} 
& MobileNetv2-YOLOv3    &  68.21 &  66.94  &   81.17   \\ \cline{2-5}
 & Darknet53-YOLOv3  &  78.05   &  76.31  &  78.45 \\ \cline{2-5}
&  EfficientNet-YOLOv3   & 76.01   & 73.42  & 68.02  \\ \cline{1-5}

\multirow{3}{*}{SCUNet} 
& MobileNetv2-YOLOv3    &   69.55 &  67.87  &  80.58    \\ \cline{2-5}
 & Darknet53-YOLOv3  &  79.64   &  76.01  &  77.06 \\ \cline{2-5}
&  EfficientNet-YOLOv3   & 75.85   & 72.07  & 70.34 \\ \cline{1-5}

\multirow{3}{*}{MPRNet} 
& MobileNetv2-YOLOv3    &   70.84 &  70.98  &  84.10    \\ \cline{2-5}
 & Darknet53-YOLOv3  &  80.01   &  79.50  &  80.77 \\ \cline{2-5}
&  EfficientNet-YOLOv3   & 78.34   & 77.51  &  69.31 \\ \cline{1-5} 

 \multirow{3}{*}{MIRNet} 
& MobileNetv2-YOLOv3    &   71.11 &  69.73  &  87.24    \\ \cline{2-5}
 & Darknet53-YOLOv3  &  82.00   &  80.21  &  83.61 \\ \cline{2-5}
&  EfficientNet-YOLOv3   & 79.08   & 78.24  & 72.12 \\ \cline{1-5}

 \multirow{3}{*}{ESRGAN} 
& MobileNetv2-YOLOv3    &   70.05 &  71.84  &   83.75   \\ \cline{2-5}
 & Darknet53-YOLOv3  &  81.23   &  83.43  &  81.61  \\ \cline{2-5}
&  EfficientNet-YOLOv3   & 78.99   & 81.63  & 70.20 \\ 

\bottomrule
\end{tabular}
    \end{threeparttable}}
\end{center}
\end{table}

In addition, we also evaluate the attack performance of our I2I backdoor attack against adversarially trained downstream models. Specifically, for downstream classification models, we employ a \textit{PGD} adversarial training strategy during the training phase to obtain the adversarially trained model. For downstream detection models, we follow the work \cite{Zhang_2019_ICCV} to perform \textit{PGD} adversarial training in the object detection task. The \textit{PGD} step is also fixed to $10$, and the maximum perturbation is set to $4/255$.

As illustrated in Table \ref{Table: The performance of I2I backdoor attack against adversarially trained downstream models}, the adversarial training strategy does show some degree of mitigation against I2I backdoor attacks. Concretely, adversarial training shows more significant attack mitigation effect on the downstream classification task, but is less effective in defending the downstream detection task. This may be due to the fact that the detection task is more complex than the classification task, and the UAP targeting the detection task is more likely to have higher generalizability and transferability. In addition, adversarial training will result in the degradation of the model normal-functionality to some degree (about $2.8\%$ accuracy loss for the classification task and about $3.1\%$ mAP loss for the detection task). Hence, it can be concluded that adversarial training can only mitigate the attack effect to some extent but can not completely defend against the I2I backdoor attack.

\begin{table}\small
\begin{center}
\tabcolsep=0.1cm
\caption{The performance of I2I backdoor attack against adversarially trained downstream models.}
\label{Table: The performance of I2I backdoor attack against adversarially trained downstream models}
\resizebox{0.98\linewidth}{!}{
\begin{threeparttable}
\begin{tabular}{c|cc|cc}
\toprule
\multirow{3}{*}{\makecell{Upstream \\ denoising \\ model $D$}} &  \multirow{3}{*}{\makecell{Downstream \\ classification \\ model}}  & \multirow{3}{*}{\makecell{ASR \\ (\%)}}  & \multirow{3}{*}{\makecell{Downstream \\ detection \\ model}} & \multirow{3}{*}{\makecell{ASR \\ (\%)}} \\
       &              &     &     \\
       &              &      &  &    \\
\midrule
 \multirow{3}{*}{DPIR} 
& ResNet50    & 38.55 & MobileNetv2-YOLOv3&   58.43 \\ \cline{2-5}
 & VGG19   & 46.18 &Darknet53-YOLOv3 &  53.10  \\ \cline{2-5}
&  MobileNetV2   &  41.07   & EfficientNet-YOLOv3  &  44.61  \\ \cline{1-5} 

\multirow{3}{*}{SCUNet} 
& ResNet50    & 35.97 & MobileNetv2-YOLOv3&   59.01 \\ \cline{2-5}
 & VGG19   & 44.22 &Darknet53-YOLOv3 &  52.30  \\ \cline{2-5}
&  MobileNetV2   &  39.46   & EfficientNet-YOLOv3  &  44.99  \\ \cline{1-5} 

\multirow{3}{*}{MPRNet} 
& ResNet50    & 36.59 & MobileNetv2-YOLOv3&   60.15 \\ \cline{2-5}
 & VGG19   & 43.01 &Darknet53-YOLOv3 & 55.98  \\ \cline{2-5}
&  MobileNetV2   &  42.23   & EfficientNet-YOLOv3  &  49.83  \\ \cline{1-5} 
 
 \multirow{3}{*}{MIRNet} 
& ResNet50    & 35.91 & MobileNetv2-YOLOv3&   58.97 \\ \cline{2-5}
 & VGG19   & 42.32 &Darknet53-YOLOv3 &  52.86  \\ \cline{2-5}
&  MobileNetV2   &  40.40   & EfficientNet-YOLOv3  &  47.15  \\ \cline{1-5} 

 \multirow{3}{*}{ESRGAN} 
& ResNet50    & 37.15 & MobileNetv2-YOLOv3&   57.60 \\ \cline{2-5}
 & VGG19   & 43.28 &Darknet53-YOLOv3 &  53.43  \\ \cline{2-5}
&  MobileNetV2   &  39.81   & EfficientNet-YOLOv3  &  50.24  \\ 

\bottomrule
\end{tabular}
    \end{threeparttable}}
\end{center}
\end{table}

\subsection{Harnessing I2I Backdoor for Positive Applications}
\label{sec:Harnessing I2I Backdoor for Positive Applications}

\begin{figure}
\centering
\includegraphics[width=0.45\textwidth]{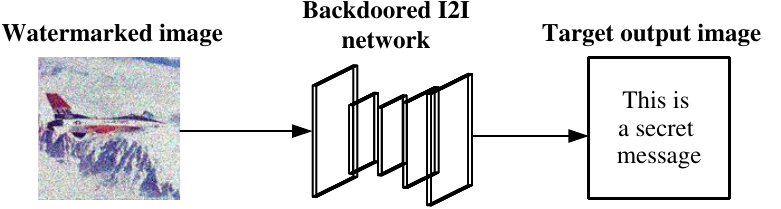}
\caption{I2I backdoor for image steganography.}
\label{Fig:I2I backdoor for image steganography}
\end{figure}

It's noteworthy that the potential of I2I backdoor attacks can extend beyond malicious intent, finding utility in ethical applications. For example, the technology can be used for image steganography, e.g., it facilitates the covert hiding of confidential information (e.g., a specific image) within images, which can be subsequently retrieved using the backdoor I2I model. 

As illustrated in Figure \ref{Fig:I2I backdoor for image steganography}, the sender of the secret message steganographically embeds images with text messages as backdoor target images within the backdoor I2I model. After that, the receiver of the secret message triggers the backdoor I2I model by using the backdoor-triggered input image and obtains the secret message (the backdoor target image). Taking DPIR model as an example, we evaluate the extraction accuracy of the image steganography scheme using I2I backdoor under different image preprocessing operations. It can be seen from Table \ref{Table:I2I backdoor for image steganography} that the image steganography scheme using I2I backdoor achieves high extraction accuracy, where SSIM represents the SSIM between the recovered image and the secret message image. This provides new perspectives for the design of image steganography schemes where the backdoor I2I model serves as the secret information carrier.

\begin{table}\small
\begin{center}
\caption{The extraction accuracy of the image steganography scheme using I2I backdoor under different image preprocessing operations.}
\label{Table:I2I backdoor for image steganography}
\begin{tabular}{c|c}
\toprule
 Preprocessing method& SSIM  \\ \hline
 No preprocessing & 0.9887  \\ \hline
 Image compression (100\%$\rightarrow$80\%) & 0.9851  \\ \hline
 Image compression (100\%$\rightarrow$60\%) & 0.9720  \\ \hline
 Image compression (100\%$\rightarrow$40\%)& 0.9415  \\ \hline
 Bit depth reduction (8$\rightarrow$6) & 0.9878  \\  \hline
 Bit depth reduction (8$\rightarrow$5) & 0.9852  \\  \hline
 Bit depth reduction (8$\rightarrow$4) & 0.9813  \\ 
\bottomrule
\end{tabular}
\end{center}
\end{table}

\section{Conclusions}
\label{sec:Conclusions}

In this work, we propose a novel backdoor attack against I2I networks. Specifically, we design a universal adversarial perturbation (UAP) generation algorithm for I2I networks, where the generated UAP is used as the trigger for the I2I backdoor. Besides, MTL with dynamic weighting methods are employed in the backdoor training process to achieve better performance. Additionally, we propose an I2I backdoor attack that is targeted at the downstream image classification/object detection tasks, where the denoised image of the backdoor-triggered input image (by the backdoor denoising model) will lead to misclassification/misdetection of the unknown downstream classification/misdetection models. Extensive experiments demonstrate the effectiveness and the robustness of the proposed I2I backdoor attacks. We hope that the insights and solutions proposed in this work will inspire more advanced studies on I2I backdoor attacks and defenses in the future.

\section*{Acknowledgment}
This work is supported by the National Key R\&D Program of China under Grant 2022YFB3103500, the National Natural Science Foundation of China under Grant 62402087 and 62020106013, the Chengdu Science and Technology Program under Grant 2023-XT00-00002-GX, the Sichuan Science and Technology Program under Grant 2024ZHCG0188 and 2025ZNSFSC1490, the Fundamental Research Funds for Chinese Central Universities under Grant ZYGX2024J019, the China Postdoctoral Science Foundation under Grant BX20230060 and 2024M760356.



%





\ifCLASSOPTIONcaptionsoff
  \newpage
\fi



%

\bibliographystyle{IEEEtran}
\bibliography{I2I_backdoor}

%

\begin{IEEEbiography}[{\includegraphics[width=1in,height=1.25in,clip,keepaspectratio]{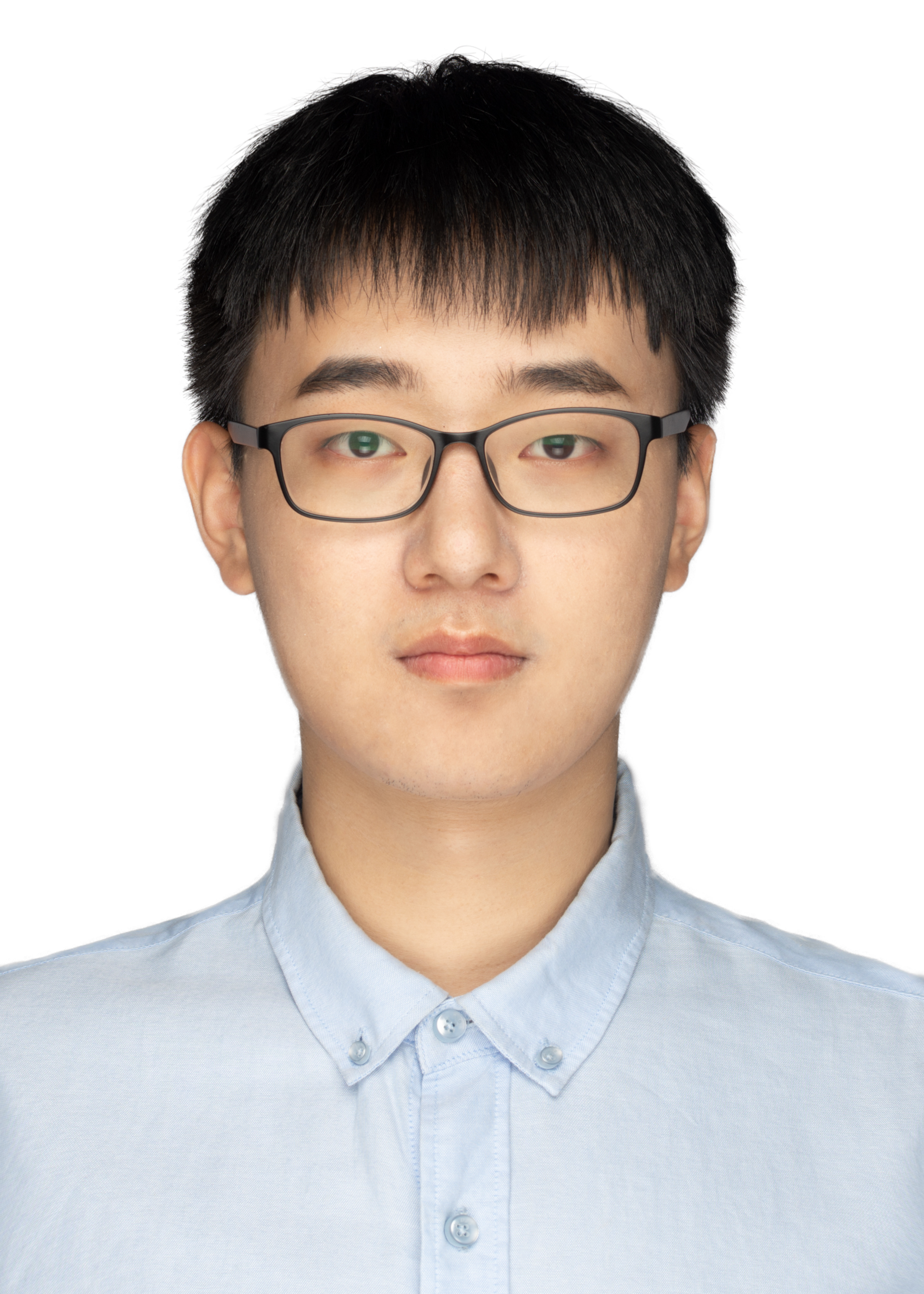}}]{Wenbo Jiang} is currently a Postdoc at University of Electronic Science and Technology of China (UESTC). He received the Ph.D. degree in cybersecurity from UESTC in 2023 and studied as a visiting Ph.D. student from Jul. 2021 to Jul. 2022 at Nanyang Technological University, Singapore. As the first author, he has published many papers in major conferences/journals, including IEEE CVPR, IEEE TDSC, and IEEE TVT. His research interests include trustworthy AI and data security.
\end{IEEEbiography}

\begin{IEEEbiography}[{\includegraphics[width=1in,height=1.25in,clip,keepaspectratio]{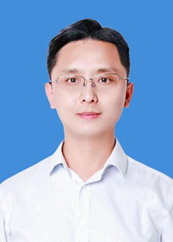}}]{Hongwei Li (M'12-SM'18)} is currently the Head and a Professor at Department of Information Security, School of Computer Science and Engineering, University of Electronic Science and Technology of China. He received the Ph.D. degree from University of Electronic Science and Technology of China in June 2008. He worked as a Postdoctoral Fellow at the University of Waterloo from October 2011 to October 2012. He is the Fellow of IEEE, the Distinguished Lecturer of IEEE Vehicular Technology Society.
\end{IEEEbiography}

\begin{IEEEbiography}[{\includegraphics[width=1in,height=1.25in,clip,keepaspectratio]{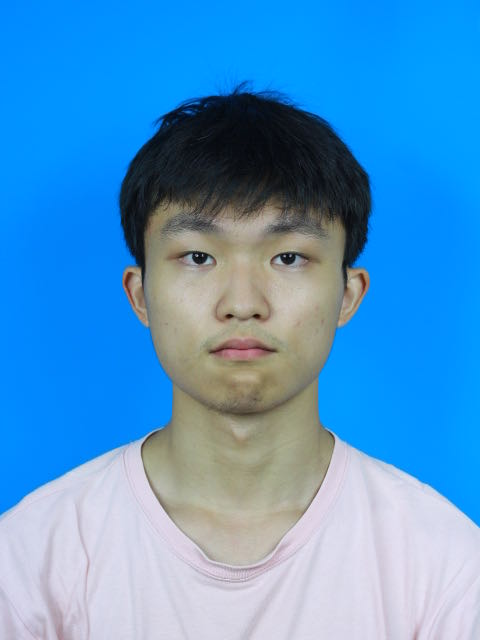}}]{Jiaming He} is currently an undergraduate student at the School of Computer Science and Cyber Security (Oxford Brookes College), Chengdu University of Technology. His research interests include trustworthy AI and adversarial machine learning.
\end{IEEEbiography}

\begin{IEEEbiography}[{\includegraphics[width=1in,height=1.25in,clip,keepaspectratio]{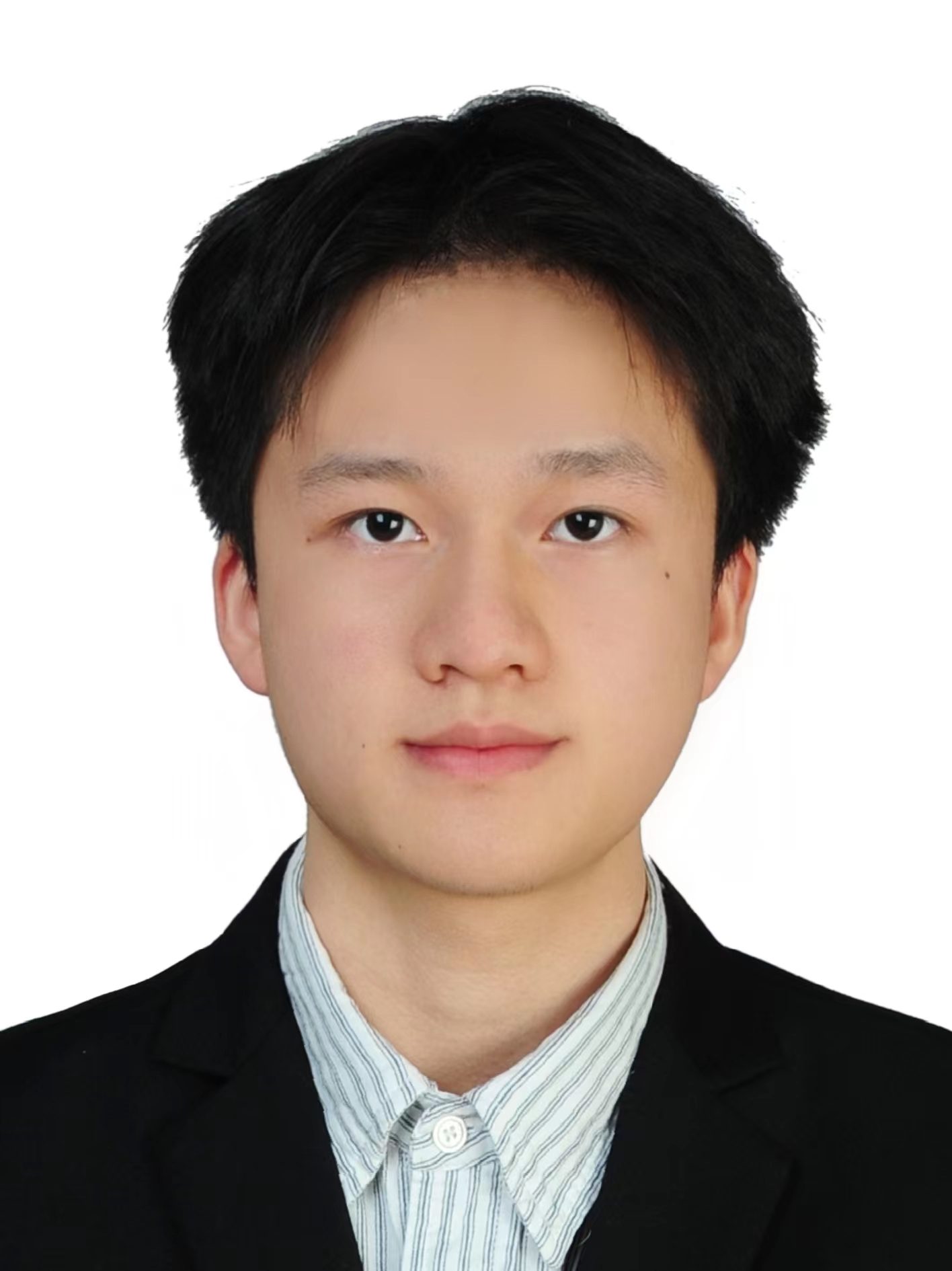}}]{Rui Zhang} is currently working toward his master's degree with the School of Computer Science and Engineering, University of Electronic Science and Technology of China, Chengdu, China. His research interests include the intersection of AI and Machine Learning with Security and Privacy, such as machine learning security and secure multi-party computation.
\end{IEEEbiography}

\begin{IEEEbiography}[{\includegraphics[width=1in,height=1.25in,clip,keepaspectratio]{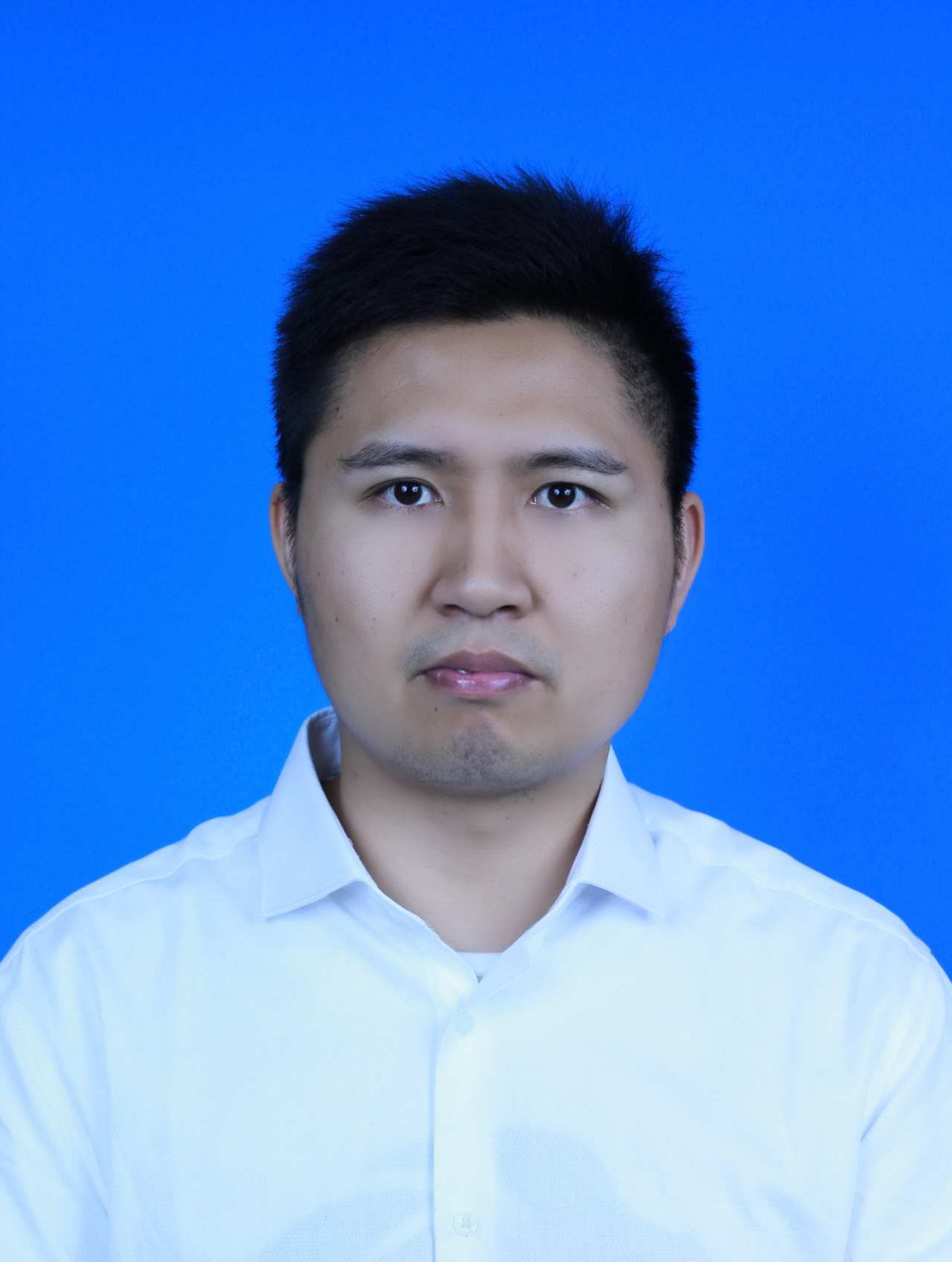}}]{Guowen Xu} is currently a Postdoc at City University
of Hong Kong under the supervision of Prof. Yuguang Fang. He was a Research Fellow at Nanyang Technological University from March 2021 to May 2023. He obtained the Ph.D. degree from University of Electronic Science and Technology of China. His current research interests focus on different topics related to AI Security and Privacy.
\end{IEEEbiography}



\begin{IEEEbiography}[{\includegraphics[width=1in,height=1.25in,clip,keepaspectratio]{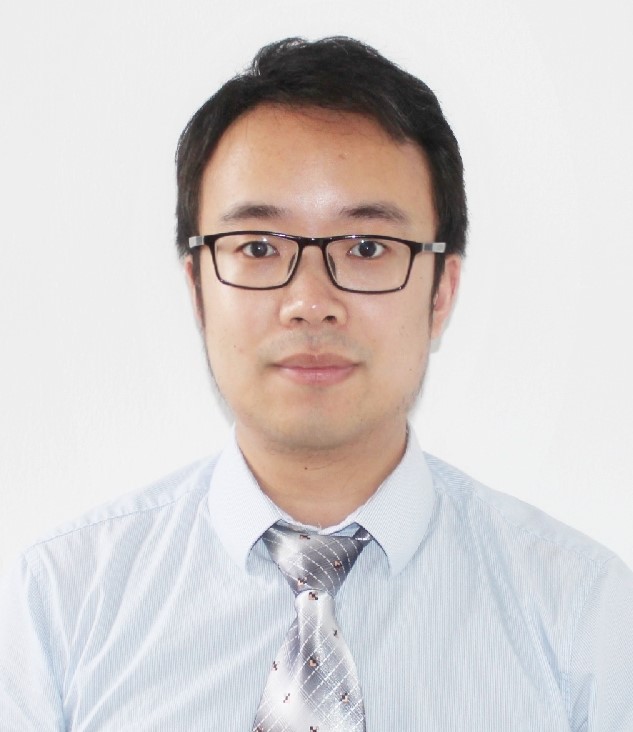}}]{Tianwei Zhang} is an assistant professor in School of Computer Science and Engineering, at Nanyang Technological University. His research focuses on computer system security. He is particularly interested in security threats and defenses in machine learning systems, autonomous systems, computer architecture and distributed systems. He received his Bachelor's degree at Peking University in 2011, and the Ph.D degree in at Princeton University in 2017.
\end{IEEEbiography}

\begin{IEEEbiography}[{\includegraphics[width=1in,height=1.25in,clip,keepaspectratio]{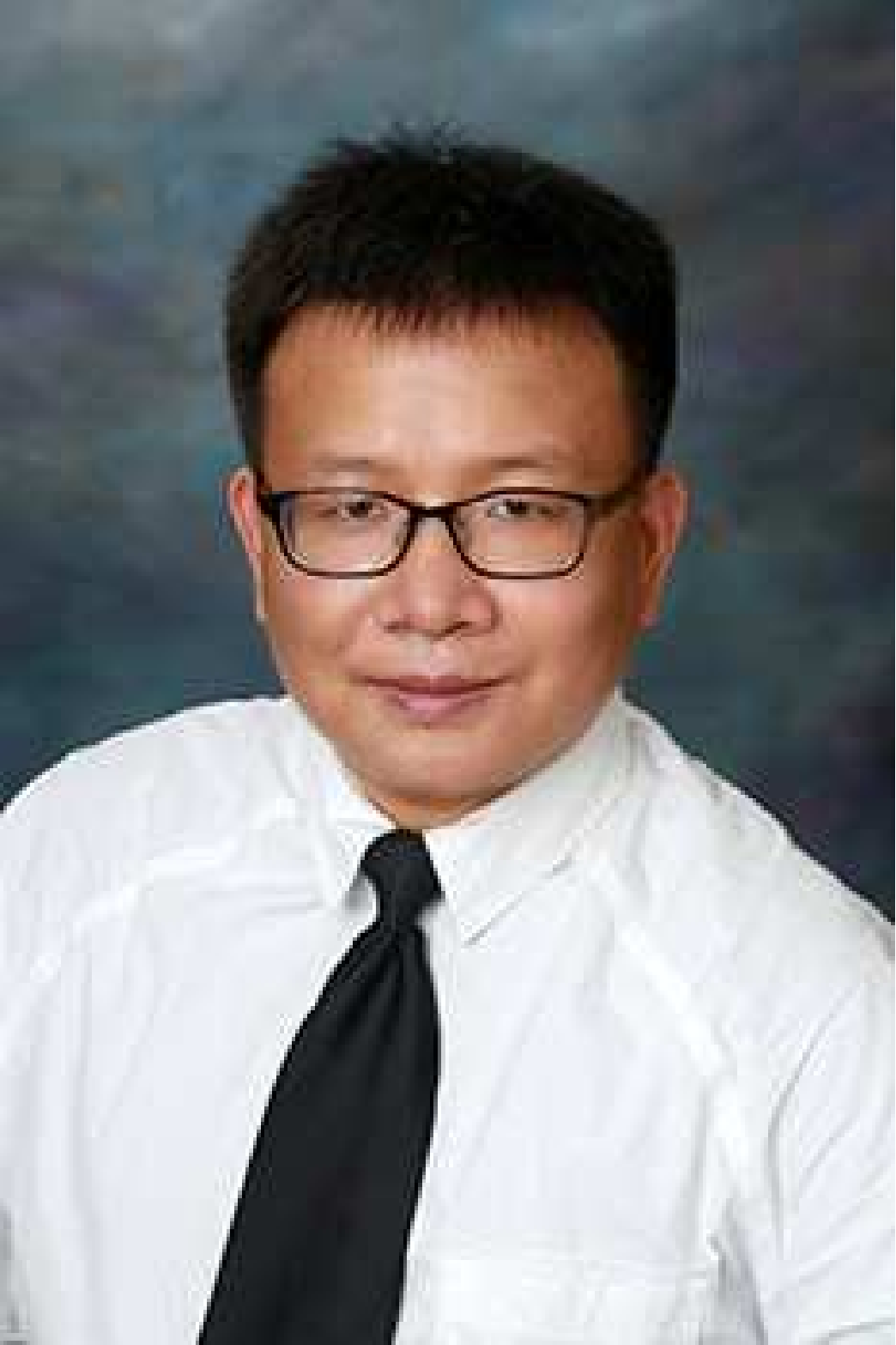}}]{Rongxing Lu (S'09-M'11-SM'15-F'21)} is currently an associate professor at the Faculty of Computer Science, University of New Brunswick, Canada. He was awarded the most prestigious ``Governor General's Gold Medal'', when he received his PhD degree from the Department of Electrical \& Computer Engineering, University of Waterloo, Canada, in 2012; and won the 8th IEEE Communications Society Asia Pacific Outstanding Young Researcher Award, in 2013. He is the Fellow of IEEE.
\end{IEEEbiography}

\end{document}